\documentclass[10pt,journal,compsoc]{IEEEtran}

\usepackage{tikz,pgf}

\usepackage[utf8]{inputenc}

\usepackage{pgfplots}
\usepackage{multirow}
\usepackage{amsmath,graphicx}
 
\usepackage[linesnumbered,ruled]{algorithm2e}
\usepackage{color}
\usepackage{float}
\restylefloat{table}
\usepackage{qtree}
\usepackage{amsfonts}       
\usepackage{nicefrac,amssymb}       
\usepackage[normalem]{ulem}
\usepackage{forest}

\definecolor{color1}{rgb}{1,0.347058823529412,0}
\definecolor{color0}{rgb}{0.862745098039216,0.0784313725490196,0.235294117647059}
\definecolor{color3}{rgb}{0.254901960784314,0.411764705882353,0.882352941176471}
\definecolor{color2}{rgb}{1,0,1}
\definecolor{color5}{rgb}{0.603921568627451,0.803921568627451,0.196078431372549}
\definecolor{color4}{rgb}{0,1,1}

\ifCLASSOPTIONcompsoc

\usepackage[nocompress]{cite}
\else

\usepackage{cite}
\fi

\ifCLASSINFOpdf

\else

\fi

\begin{document}

\title{Minimum Uncertainty Based Detection of Adversaries in Deep Neural Networks}
\author{Fatemeh Sheikholeslami,~\IEEEmembership{Student Member,~IEEE,}
        Swayambhoo Jain,~\IEEEmembership{Member ,~IEEE,\\}
        and Georgios B. Giannakis,~\IEEEmembership{Fellow,~IEEE}
\IEEEcompsocitemizethanks{\IEEEcompsocthanksitem Part of this work was done during a summer research internship at Technicolor AI Lab in Palo Alto, CA - USA. This research was supported in part by NSF grant 151405\textbackslash6, 1505970, 1901134, and 1711471.  Author emails: \texttt{sheik081@umn.edu, swayambhoo.jain@gmail.com, georgios@umn.edu }
}
}

\markboth{Journal of \LaTeX\ Class Files,~Vol.~?, No.~?, ~?}%
{Shell \MakeLowercase{\textit{et al.}}: Bare Demo of IEEEtran.cls for Computer Society Journals}

\IEEEtitleabstractindextext{%
	
\begin{abstract}
Despite their unprecedented performance in various domains, utilization of Deep Neural Networks (DNNs) in safety-critical environments is severely limited in the presence of even small adversarial perturbations. The present work develops a randomized approach to detecting such perturbations based on minimum uncertainty metrics that rely on sampling at the hidden layers during the DNN inference stage. { Inspired by Bayesian approaches to uncertainty estimation,} the sampling probabilities are designed for effective detection of the adversarially corrupted inputs. Being modular, the novel detector of adversaries can be conveniently employed by any pre-trained DNN at no extra training overhead. Selecting which units to sample per hidden layer entails quantifying the amount of DNN output uncertainty, where the overall uncertainty is expressed in terms of its layer-wise components - what also promotes  scalability. Sampling probabilities are then sought by minimizing uncertainty measures
layer-by-layer, leading to a novel convex optimization problem that admits an exact solver with superlinear convergence rate. By simplifying the objective function, low-complexity approximate solvers are also developed. In addition to valuable insights, these approximations link the novel approach with state-of-the-art randomized adversarial detectors. The effectiveness of the novel detectors in the context of competing alternatives is highlighted through extensive tests for various types of adversarial attacks with variable levels of strength.
\end{abstract}

\begin{IEEEkeywords}
Adversarial input, Bayesian neural networks, attack detection, uncertainty estimation.
\end{IEEEkeywords}}

\maketitle

\IEEEdisplaynontitleabstractindextext

\IEEEpeerreviewmaketitle

\IEEEraisesectionheading{\section{Introduction}\label{sec:intro}} 
Unprecedented learning capability offered by Deep Neural Networks (DNNs) has enabled state-of-the-art performance in diverse tasks such as object recognition and detection \cite{simonyan2014very, redmon2017yolo9000, sharif2016accessorize}, speech recognition and language translation \cite{sutskever2014sequence}, voice synthesis \cite{engel2017neural}, and many more, to reach or even surpass human-level accuracy. Despite their performance however, recent studies have cast doubt on the reliability of DNNs as highly-accurate networks are shown to be extremely sensitive to carefully crafted inputs designed to fool them \cite{szegedy2013intriguing, moosavi2017universal, tsipras2018there}. Such fragility can easily lead to sabotage once adversarial entities target critical environments such as autonomous cars \cite{eykholt2018robust}, automatic speech recognition \cite{zhang2017dolphinattack}, and  face detection \cite{goswami2018unravelling, bose2018adversarial, sharif2016accessorize}. The extreme brittleness of convolutional neural networks (CNNs) for image classification is highlighted since small adversarial perturbations on the clean image, although often imperceptible to the human eye, can lead the trained CNNs to classify the \textit{adversarial examples} incorrectly with high confidence. In particular, design of powerful adversarial perturbations in environments with different levels of complexity and knowledge about the target CNN, known as white, grey, and black-box attacks, have been investigated in several works \cite{ moosavi2017universal, carlini2017adversarial, carlini2017towards, papernot2016transferability,zhang2019limitations,hong2019zero}. These considerations motivate well the need for designing robust and powerful attack detection mechanisms for reliable and safe utilization of DNNs~\cite{papernot2016limitations}. 

Defense against adversarial perturbations has been mainly pursued in two broad directions: (i) attack \emph{detection}, and (ii) attack \emph{recovery}. Methods in the first category aim at detecting adversarial corruption in the input by classifying the input as clean or adversarial, based on tools as diverse as auto-encoders \cite{gu2014towards}, detection sub-networks  \cite{metzen2017detecting, lu2017safetynet}, and dropout units \cite{feinman2017detecting}. On the other hand, methods in the second category are based on recovery schemes that robustify the classification by data pre-processing and randomization \cite{random_certified, guo2017countering, gopalakrishnan2018combating}, adversarial training \cite{miyato2018virtual, schmidt2018adversarially, sinha2017certifying, EricZico, towards_2019}, sparsification of the network \cite{NIPS2018_sparse, targetted_dropout} and Lipschitz regularization \cite{zhang2016understanding, gouk2018regularisation}, to name just a few. 

Furthermore, the so-termed {\emph{over-confidence}} of DNNs in classifying ``out-of-distribution,'' meaning samples which lie in unexplored regions of the input domain, or even ``misclassified'' samples, has been unraveled in \cite{hendrycks17baseline, calibration}. This has motivated the need for uncertainty estimation as well as calibration of the networks for robust classification. Modern Bayesian neural networks target this issue by modeling the distribution of DNN weights as random \cite{malinin2018predictive}, and estimating the  DNN output uncertainty through predictive entropy, variance, or mutual information \cite{smith2018understanding, kendall2017uncertainties, feinman2017detecting, icassp}. The well-known {dropout} regularization technique is one such approximate Bayesian neural network, now widely used in training and testing of DNNs \cite{gal2016bayesian, gal2016theoretically, Statistical_2019}. 
 
Moreover, approaches relying on dropout units have shown promising performance in successfully detecting adversarial attacks, where other defense mechanisms fail \cite{carlini2017adversarial}. In particular, \cite{feinman2017detecting} utilizes randomness of dropout units during the test phase as a defense mechanism, and approximates the classification uncertainty by Monte Carlo (MC) estimation of the output variance. Based on the latter, images with high classification uncertainty are declared as adversarial. Recently, dropout defense has been generalized to non-uniform sampling \cite{dhillon2018stochastic}, where entries of the hidden-layers are randomly sampled, with probabilities proportional to the entry values. This heuristic sampling of units per layer is inspired by intuitive reasoning: activation units with large entries have more information and should be sampled more often~\cite{dhillon2018stochastic}. However, analytical understanding  has not been investigated. 

The goal here is to further expand the understanding of uncertainty estimation in DNNs, and thereby improve the detection of adversarial inputs. The premise is that inherent distance of the adversarial perturbation from the natural-image manifold will cause the overall network uncertainty to exceed that of the clean image, and thus successful detection can be obtained. 

To this end, and inspired by \cite{dhillon2018stochastic}, we rely on random sampling of units per hidden layer of a pre-trained network to introduce randomness. Moreover, {{ inspired by the Bayesian approaches to uncertainty estimation}}, the overall uncertainty of a given image is then quantified in terms of its hidden-layer components. We then formulate the task of adversary detection as uncertainty minimization by optimizing over the sampling probabilities to provide effective detection. Subsequently, we develop an exact solver with super-linear convergence rate as well as approximate low-complexity solvers for an efficient layer-by-layer uncertainty minimization scheme. Furthermore, we draw connections with uniform dropout \cite{feinman2017detecting} as well as stochastic approximate pruning (SAP) \cite{dhillon2018stochastic}, and provide an efficient implementation of the novel approach by interpreting it as a non-uniform dropout scheme. Extensive numerical tests on CIFAR10 and high-quality cats-and-dogs images in the presence of various attack schemes corroborate the importance of our designs of sampling probabilities, as well as the placement of sampling units per hidden layer for improved detection of adversarial inputs.

The rest of the paper is organized as follows. An overview on Bayesian inference and { {uncertainty-based}} detection in neural networks is provided in Section \ref{sec:bayesian_intro}. {Inspired by this}, the proposed class of detectors is introduced in Section \ref{sec:UncertaintyMinimization}, and exact as well as low-complexity approximate solvers for the layer-by-layer uncertainty minimization are the subjects of Section \ref{sec:solvers}. Implementation issues are dealt with in Section \ref{sec:implementation}, numerical tests are provided in Section \ref{sec:tests}, and concluding remarks are discussed in Section \ref{sec:conclusion}.

\section{Bayesian Neural Network Preliminaries} \label{sec:bayesian_intro}
Bayesian inference is among the powerful tools utilized for analytically understanding and quantifying uncertainty in DNNs \cite{srivastava2014dropout ,gal2016theoretically}.
In this section, we  provide a short review on the basics of Bayesian neural networks, and move on to the inference phase for adversary detection in Section \ref{sec:my_detection}, which is of primary interest in this work. 

Consider an $L$-layer deep neural network, which maps the input $\mathbf{x} \in \mathcal{X}$ to output $\mathbf{y} \in \mathcal{Y}$. The weights are denoted by $\omega := \{\mathbf{ W}_l\}_{l=1}^L$, and are modeled as random variables with prior probability density function (pdf) $p(\omega)$. 

Given training input $\mathbf{X} := [\mathbf{x}_1, \mathbf{x}_2, ..., \mathbf{ x}_n]$ and output data $\mathbf{Y} := [\mathbf{y}_1, \mathbf{y}_2, ..., \mathbf{y}_n]$, it is assumed that the parameters $\omega$ only depend on these $(\mathbf{X},\mathbf{Y})$ data.  As a result, the predictive pdf for a new input $\mathbf{ x}_\nu$ can be obtained via marginalization as \cite{gal2016bayesian}
\begin{equation} \label{predictive}
p(\mathbf{y}_\nu|\mathbf{ x}_\nu, \mathbf{X}, \mathbf{Y }) = \int p(\mathbf{y}_\nu|\mathbf{ x}_\nu, \omega) p(\omega| \mathbf{X}, \mathbf{Y}) d\omega
\end{equation}
which requires knowing the conditional $p(\omega| \mathbf{X}, \mathbf{Y})$. The complexity of estimating $p(\omega| \mathbf{X}, \mathbf{Y})$ motivates well the variational inference (VI) approach, where $p(\omega| \mathbf{X}, \mathbf{Y})$ is replaced by a surrogate pdf  $q_\theta(\omega)$ that is parameterized by $\theta$. For $q_\theta(\omega)$, it is desired to: (D1) approximate closely $p(\omega| \mathbf{X}, \mathbf{Y})$; and, (D2) provide easy marginalization in \eqref{predictive} either in closed form or empirically. To meet (D1), the surrogate is chosen by minimizing the Kullback-Leibler (KL) divergence $KL(p(\omega| \mathbf{X}, \mathbf{Y}), q_\theta(\omega))$, which is subsequently approximated by the log evidence lower bound \cite[p.~462]{bishop}
\begin{equation}\label{log_evidence}
\mathcal{L}_{VI} (\theta) := \int q_\theta(\omega) \log p(\mathbf{Y}| \mathbf{X}, \omega) d \omega - KL(q_\theta(\omega)||p(\omega))\;.
\end{equation}
Finding $q_\theta$ boils down to maximizing the log evidence lower bound, that is, $\theta_{\text{VI}} = \arg \max_\theta \mathcal{L}_{\text{VI}} (\theta)$. A common choice for $q_\theta(\omega)$ to also satisfy (D2) is described next.

\subsection{Variational inference} \label{sec:vi}
A simple yet effective choice for $q_\theta(\omega)$ is a factored form modeling the weights as independent across layers,  that is
\begin{equation}\label{q}
q_{{\theta}}(\omega) = \prod_{l=1}^{L} q (\mathbf{W}_l; \mathbf{M}_l,\boldsymbol{\theta}_{z_l})
\end{equation}
where the $l$-th layer with $h_{l}$ hidden units is modeled as 
\begin{equation}\label{W}
\mathbf{W}_l = \mathbf{M}_l \text{diag}([z_{l,1}, z_{l,2}, \ldots, z_{l,h_l}])\:, \quad  l=1,\ldots,L
\end{equation} 
where $\mathbf{M}_l$ is an $h_{l+1} \times h_{l}$ deterministic weight matrix multiplied by a diagonal matrix formed by the binary random vector $\mathbf{z}_l := [z_{l,1}, z_{l,2}, \ldots, z_{l,h_l}] \in \{0,1\}^{h_l}$ with entries drawn from a pmf $q_z(\mathbf{z}_l;\boldsymbol{\theta}_{z_l})$ parameterized by $\boldsymbol{\theta}_{z_l}$. 

If the entries $\{z_{l,i}\}$ are i.i.d. Bernoulli with (identical) probability (w.p.) $\pi$, they effect what is referred to as \emph{uniform} (across layers and nodes) \emph{dropout}, which is known to prevent overfitting \cite{srivastava2014dropout}. Clearly, the parameter set ${\theta} := \{\mathbf{M}_l, \boldsymbol{\theta}_{z_l}\}_{l=1}^L = \{\mathbf{M}_l \}_{l=1}^L \cup \{\pi\}$ fully characterizes $q_\theta(\omega)$. The dropout probability $1-\pi$ is preselected in practice, while $\{\mathbf{M}_l\}_{l=1}^L$ can be obtained using the training data by maximizing the log evidence lower bound in \eqref{log_evidence}. Nonetheless, integration in \eqref{log_evidence} over all the Bernoulli variables is analytically challenging, while sampling from the Bernoulli pmf is relatively cheap. This prompts approximate yet efficient integration using Monte Carlo estimation. A more detailed account of training Bayesian neural networks can be found in~\cite{gal2016bayesian, bishop, neal2012bayesian}. Moving on, the ensuing subsection deals with detection of adversarial inputs { inspired by Bayesian neural networks}.

\subsection{Detection of DNN adversaries}\label{sec:my_detection}
{In addition to facilitating ELBO maximization during the training phase, probabilistic view on the network parameters during the test phase can also be utilized towards output uncertainty estimation for the detection of adversarial inputs\cite{gal2016bayesian}. To do so, detection during the testing phase proceeds by approximating the predictive pdf in \eqref{predictive} using the variational surrogate $q_\theta(\omega)$, as}
%
\begin{equation}\label{pred}
p(\mathbf{y}_\nu|\mathbf{ x}_\nu, \mathbf{X}, \mathbf{Y }) \approx \int p(\mathbf{y}_\nu|\mathbf{ x}_\nu, \omega) q_\theta(\omega) d\omega.
\end{equation}
Deciphering whether a given input $\mathbf{x}_\nu$ is adversarial entails three steps: (S1) parametric modeling of $q_\theta(\omega)$; (S2) estimating the DNN output \emph{uncertainty} captured by $p(\mathbf{y}_\nu|\mathbf{ x}_\nu, \mathbf{X}, \mathbf{Y })$; and (S3) declaring $\mathbf{x}_\nu$ as adversarial if the output uncertainty exceeds a certain threshold, and clean otherwise. These steps are elaborated next.

\textit{Step 1: Parametric modeling of $q_\theta{(\omega)}$.} Recall that uniform dropout offers a popular special class of $q_\theta(\omega)$ pdfs, and has been employed in adversary detection \cite{feinman2017detecting}. Here, we specify the richer model of $q_\theta(\omega)$ in \eqref{q} and \eqref{W} that will turn out to markedly improve detection performance. Different from uniform dropout, we will allow for (possibly correlated) Bernoulli variables with carefully selected (possibly non-identical) parameters. If such general $\{\boldsymbol{\theta}_{z_l}\}_{l=1}^L$ can be obtained, matrices $\{\mathbf{M}_l\}_{l=1}^L$ are then found as follows. 

Let $\{\mathbf{W}^{(TR)}_l\}_{l=1}^L$ be deterministic weight matrices obtained via non-Bayesian training that we denote as (TR)\footnote{Such as back propagation based on e.g., a cross-entropy criterion.}. We will use  $\mathbf{W}^{(TR)}_l$ to specify the mean of the random weight matrix $\mathbf{W}_l$ in our approach, meaning we choose $\mathbb{E}_{q_{z}(\mathbf{z}_l; \boldsymbol{\theta}_l)} [\mathbf{W}_l]\mathbf{x}_{(l-1)} = \mathbf{W}^{(TR)}_l \mathbf{x}_{(l-1)}$ $\forall l$, where $\mathbf{x}_{(l-1)}$ is the output of the $(l-1)$st layer for a given input $\mathbf{x}_\nu$ passing through the DNN with deterministic weights $\{\mathbf{W}^{(TR)}_l\}_{l=1}^L$. With 
$\mathbf{W}^{(TR)}_l$ available, we first design $q_z(\mathbf{z}_l;\boldsymbol{\theta}_{z_l})$; next, we find $\boldsymbol{\theta}_{z_l}$; and then $\mathbf{M}_l$, as 
\begin{equation}\label{M2}
	\mathbf{M}_l = \mathbf{W}^{(TR)}_l \; \text{diag}^{\dagger}\Big(\mathbb{E}_{q_z(\mathbf{z}_l; \boldsymbol{\theta}_{z_l})} [\mathbf{z}_{l}] \Big)\:, \,\;\;l=1,\ldots, L
\end{equation}
where the pseudo-inverse $\dagger$ means that inverse entries are replaced with zeros if $\mathbb{E}_{q_z(\mathbf{z}_l; \boldsymbol{\theta}_{z_l})} [z_{l,i}] = 0$. 

\emph{Step 2: Quantifying the DNN output uncertainty.} 
Since evaluation of $p(\mathbf{y}_\nu|\mathbf{ x}_\nu, \mathbf{X}, \mathbf{Y })$ in \eqref{pred} is prohibitive, one can estimate it using MC sampling. In particular, one can readily obtain MC estimates of (conditional) moments of $\mathbf{y}_\nu$. For instance, its mean and variance can be estimated as
\begin{equation*}
\mathbb{E}_{q_\theta(\omega)}[\mathbf{ y}_\nu| \mathbf{x}_\nu; \{\boldsymbol{\theta}_{\mathbf{z}_l}\}_{l=1}^L ] 
\simeq \bar{\mathbf{y}}_\nu = \dfrac{1}{R} \sum_{r=1}^R \mathbf{y}_{\nu}^{(r)}
\end{equation*}
and
\begin{align}
\text{Cov}_{q_\theta(\omega)}[\mathbf{ y}_\nu| \mathbf{x}_\nu; \{\boldsymbol{\theta}_{\mathbf{z}_l}\}_{l=1}^L] 
\simeq &   \dfrac{1}{R} \sum_{r=1}^R \mathbf{y}_{\nu}^{(r)}  \mathbf{y}_{\nu}^{\top(r)}  -\bar{\mathbf{y}}_\nu \bar{\mathbf{y}}_\nu^\top \label{sample_cov}
\end{align}
where $\mathbf{y}_{\nu}^{(r)}$ is the output of the $r$-th DNN realized through weights 
$\{\mathbf{W}_l^{(r)}\}_{l=1}^L$ with input $\mathbf{x}_\nu$. The predictive variance is the trace of $\text{Cov}_{q_\theta(\omega)}[\mathbf{ y}_\nu| \mathbf{x}_\nu; \{\boldsymbol{\theta}_{\mathbf{z}_l}\}_{l=1}^L]$ that we henceforth abbreviate as $\text{Tr}(\text{Cov}_{q_\theta(\omega)}[\mathbf{ y}_\nu| \mathbf{x}_\nu])$. Given $\mathbf{x}_\nu$, the latter has been used to quantify output uncertainty as 
$U(\mathbf{x}_\nu) = \text{Tr}(\text{Cov}_{q_\theta(\omega)}[\mathbf{ y}_\nu| \mathbf{x}_\nu])$~\cite{feinman2017detecting}. Additional measures of uncertainty will be presented in the next section. 

\emph{Step 3: Detecting adversarial inputs}. Given $U(\mathbf{x}_\nu)$, detection of adversarial inputs is cast as testing the hypotheses 
\begin{align}\label{test}
\begin{cases}
\mathcal{H}_0: \mathbf{x}_\nu = \mathbf{x}_\nu^{\text{clean}}\qquad \quad \qquad U(\mathbf{x}_\nu)\leq\tau_0\\
\mathcal{H}_1:\mathbf{x}_\nu = \mathbf{x}_\nu^{\text{clean}} +\mathbf{n}_\nu^{\text{adv}} \qquad U(\mathbf{x}_\nu)>\tau_0\\
\end{cases}	
\end{align}
where the null suggests absence of adversarial perturbation (low variance/uncertainty below threshold $\tau_0$), while the alternative in effect raises a red flag for presence of adversarial input (high variance/uncertainty above threshold $\tau_0$). 

We will now proceed to introduce our novel variational distribution model targeting improved detection of adversaries based on uncertainty minimization. 

\section{Minimum Uncertainty based Detection} \label{sec:UncertaintyMinimization}
To design $q_z(\mathbf{z}_l;\boldsymbol{\theta}_{z_l})$, we will build on and formalize the sampling scheme in~\cite{dhillon2018stochastic} that is employed to specify the joint pmf of the (generally correlated) binary variables $\{ z_{l,i}\}_{i=1}^{h_l}$ per layer $l$. To this end, we randomly pick one activation unit output of the $h_l$ hidden units per layer $l$; and repeat such a random draw $C$ times with replacement. Let $\boldsymbol{\zeta}_l^{(c)} $ denote 
per draw $c$ the $h_l \times 1$ vector variable 
\[\boldsymbol{\zeta}_l^{(c)} = [{\zeta}_{l,1}^{(c)}, {\zeta}_{l,2}^{(c)} \ldots  {\zeta}_{l,h_l}^{(c)}]^\top \sim {\text{Categorical}(\mathbf{ p}_l)}\:,\;\;c=1,\ldots,C\]
where each entry $\zeta_{l,i}^{(c)}$ is a binary random variable with 
\begin{align*}
{\zeta}_{l,i}^{(c)} = 
\begin{cases}
1 \qquad  \text{if draw $c$ picks the $i$th unit of hidden layer $l$} \\
0 \qquad \text{otherwise}
\end{cases}
\end{align*}
and the $h_l\times 1$ vector $\mathbf{p}_l$ with nonegative entries summing up to $1$ specifies the Categorical pmf of $\boldsymbol{\zeta}_l^{(c)}$.

With $||$ denoting element-wise binary \emph{OR} operation on vectors $\{\boldsymbol{\zeta}_{l}^{(c)}\}_{c=1}^C$, we define next the vector  
\begin{equation}\label{z_l}
\mathbf{z}_{l} := \boldsymbol{\zeta}_{l}^{(1)} \; || \; \boldsymbol{\zeta}_{l}^{(2)} \; || \;  ... \; || \; \boldsymbol{\zeta}_{l}^{(C)}\;.
\end{equation}
Using $\mathbf{z}_{l}$ as in \eqref{z_l} with  ${\{\boldsymbol{\theta}_{z_l} = \mathbf{p}_l\}_{l=1}^L}$
to be selected, enables finding the expectation and then $\mathbf{M}_l$ in \eqref{M2}. Deterministic matrix $\mathbf{M}_l$ along with the variates $\{\mathbf{z}_{l}^{(r)}\}_{r=1}^R$ provide the desired DNN realizations to estimate the uncertainty $U(\mathbf{x}_\nu;{\{\mathbf{p}_l\}_{l=1}^L}) = \text{Tr}(\text{Cov}_{q_\theta(\omega)}[\mathbf{ y}_\nu| \mathbf{x}_\nu])$ as in \eqref{sample_cov}. In turn, this 
leads to our novel adversarial input detector (cf. \eqref{test})
\begin{align}\label{test_new}
\begin{cases}
\mathcal{H}_0: \mathbf{x}_\nu = \mathbf{x}_\nu^{\text{clean}}  \;\;\;\; \min_{\{\mathbf{p}_l\}_{\forall l}} U (\mathbf{x}_\nu; \{\mathbf{p}_l\}_{l=1}^L)\leq\tau_0\\
\mathcal{H}_1:\mathbf{x}_\nu = \mathbf{x}_\nu^{\text{clean}} +\mathbf{n}_\nu^{\text{adv}}\qquad \qquad {\text{otherwise}}
\end{cases}
\end{align} 
where variational parameters ${\{\mathbf{p}_l\}_{l=1}^L}$ are sought such that  uncertainty  $U(\mathbf{x}_\nu; \{\mathbf{p}_l\}_{l=1}^L)$ is minimized under $\mathcal{H}_0$.

{
	
The rationale behind our detector in \eqref{test_new} is that for a given detection threshold $\tau_0$, uncertainty minimization will increase the number of clean images whose minimized uncertainty will fall below this threshold, and thus lead to a lower probability of false alarms. The probability of  adversarial input detection however, depends on test statistic pdf under $\mathcal{H}_1$, in which the adversarial perturbation $\mathbf{n}_\nu^{\text{adv}}$ is unknown. The premise here is that due to \textit{network instability} under $\mathcal{H}_1$, the sought probabilities  ${\{\mathbf{p}_l\}_{l=1}^L}$ will not reduce uncertainty under $\mathcal{H}_1$ \textit{as effectively}, thus minimum uncertainty-based detection can provide improved ROC curves. In lieu of analytical metrics, this has been tested through extensive numerical experiments, and its effectiveness has been empirically corroborated. }
	

{
Furthermore, Table 1 provides a list of variables and their definition to improve readability.

\begin{table}[!h]
	\centering
	\begin{tabular}{|c|l|}
		\hline
		Param. & Definition \\
		\hline
		{	$\mathbf{x}_\nu$ }& 	{Test input image }\\
		{	$\mathbf{y}_\nu^{\text{target}}$}&	{ Ground-truth one-hot class label  for input $\mathbf{x}_\nu$}\\
			{$\mathbf{y}_\nu$} & 	{Output for $\mathbf{x}_\nu$ by the deterministic classifier}\\
		{	$\mathbf{y}_\nu^{(r)}$} &	{ Output for $\mathbf{x}_\nu$ in the $r$-th realization} \\
		&	{ of the detection network (with sampling units)}\\
		{$\mathbf{x}_{(l)}$} &	{Value at hidden layer $l$ with width $h_l$}\\
		{$\mathbf{W}_l$ }& 	{Network weight at layer $l$ (random variable)} \\
		
			{$\mathbf{W}_l^{\text{TR}}$ }& 	{Network weights given by the training phase }\\
		
		{	$\mathbf{M}_l$ }&	{ Expected value of $\mathbf{W}_l$}\\
			{$\mathbf{S}_l$ }&	{ Sampling matrix, defined as  $\mathbf{S}_l = \text{diag}( \mathbf{z}_{l} )$} \\
		
		{$\mathbf{D}_l$} & 	{Diagonal matrix 
		$\mathbf{D}_l = \text{diag}^{\dagger}\Big(\mathbb{E}_{q_z(\mathbf{z}_l; \mathbf{p}_l)} [\mathbf{z}_{l}] \Big)$ }\\
		{$\mathbf{z}_l$}& 	{$h_l$-dimensional binary multivariate, modeling}  \\ &	{the overall 
		sampling outcome at layer $l$} \\
			{$\boldsymbol{\zeta}_l^c$ }& 	{Categorical $h_l$-dimensional binary multivariate, } \\ 
		&	{modeling the  $c$-th  draw at layer $l$}\\
		{$\mathbf{p}_l$ }&	{Parameters of the categorical pmf  of $\boldsymbol{\zeta}_l^{(c)}$}\\
			{$C$} & 	{Total number of draws}\\
		{$f$}&	{ Scaler coeff. in $[0,1] $ defining $C = f \times \text{nnz}(\mathbf{x}_l)$ }\\
		{$B$}& 	{Resnet architecture blocks in Tables 3 and 4}\\
		\hline
	\end{tabular}\label{tab:variables}
	{\caption{List of variables} }
\end{table}
}

\subsection{Uncertainty measures}
In order to carry the hypothesis test in \eqref{test_new}, one has options for $U(\mathbf{x}_\nu; \{\mathbf{p}_l\}_{l=1}^L)$ other than the conditional variance. For DNNs designed for classification, mutual information has been recently proposed as a measure of uncertainty  \cite{smith2018understanding}
\begin{equation}\label{mutual}
\hat{I} (\mathbf{x}_\nu; \{\mathbf{p}_l\}_{l=1}^L) := H(\bar{\mathbf{y}}_\nu) - \dfrac{1}{R} \sum_{r=1}^R H(\mathbf{y}_{\nu}^{(r)})
\end{equation}
where superscript $r$ indexes the pass of input $\mathbf{x}_\nu$ through the $r$th DNN realization with corresponding random output $\mathbf{y}_\nu^{(r)} :=[y_{\nu,1}^{(r)}, y_{\nu,2}^{(r)}, \ldots, y_{\nu,K}^{(r)}]^\top$ in a $K$-class classification task, and $H(.)$ is the entropy 
function\footnote{Entropy functions in \eqref{mutual} are also parameterized by $\{\mathbf{p}_l\}_{l=1}^L$, but we abbreviate them here as $H(\bar{\mathbf{y}}_\nu)$ and  $H(\mathbf{y}_{\nu}^{(r)})$.}  
\begin{equation}\label{entropy}
H(\mathbf{y}_\nu):= - \sum_{k=1}^K y_{\nu,k} \log(y_{\nu,k}).
\end{equation}
The test statistic in \eqref{test_new} requires finding $\{\mathbf{p}_l\}_{l=1}^L$ by solving
\begin{equation}\label{opt_mutual}
\min_{\{\mathbf{p}_l\}_{l=1}^L} \hat{I} (\mathbf{x}_\nu; \{\mathbf{p}_l\}_{l=1}^L) 
\end{equation}
which is highly non-convex. However, using Taylor's expansion of the logarithmic terms in \eqref{entropy}, one can approximate the mutual information in \eqref{mutual} with the variance score $\text{Tr}(\text{Cov}_{q_\theta(\omega)}[\mathbf{ y}_\nu])$ in \eqref{test_new}, where the conditioning on $\mathbf{x}_\nu$ has been dropped for brevity~\cite{smith2018understanding}. As a result, 
the optimization in \eqref{opt_mutual} is approximated as
\begin{equation}\label{opt_var}
\min_{\{\mathbf{p}_l\}_{l=1}^L}  U (\mathbf{x}_\nu;  \{\mathbf{p}_l\}_{l=1}^L) = \text{Tr}(\text{Cov}_{q_\theta(\omega)}[\mathbf{ y}_\nu]) \;. 
\end{equation}
To solve \eqref{opt_var}, one needs to express the objective in terms of the optimization variables ${\{\mathbf{p}_l\}} $ for all layers explicitly. To this end, the following section studies a two-layer  network, whose result will then be generalized to deeper models. 

\subsection{Simplification of the predictive variance }
Aiming at a convenient expression for the cost in \eqref{opt_var}, consider first a two-layer network with input-output (I/O) relationship\footnote{Derivations in this section carry over readily to a more general I/O $
		\mathbf{y}_\nu = \sigma_{\text{softmax}} \Big(\mathbf{W}_2 \sigma ( \mathbf{W}_1 \mathbf{x}_\nu+\mathbf{b}_1)+ \mathbf{b}_2\Big)$ with $\mathbf{b}_1$ and $\mathbf{b}_2$ deterministic.}
\begin{equation}\label{neural_network}
\mathbf{y}_\nu = \sigma_{\text{softmax}} \Big(\mathbf{W}_2 \sigma ( \mathbf{W}_1 \mathbf{x}_\nu)\Big)
\end{equation}
where $\mathbf{W}_1, \mathbf{W}_2$ are random matrices corresponding to the weights of the two layers as in \eqref{M2}, while $\sigma_{\text{softmax}}$ is the softmax memoryless nonlinearity
\[\sigma_{\text{softmax}}(\mathbf{u}) := \left[\dfrac{e^{u_1}}{\sum_{i=1}^K{e^{u_i}}}, \dfrac{e^{u_2}}{\sum_{i=1}^K{e^{u_i}}},\ldots, \dfrac{e^{u_K}}{\sum_{i=1}^K{e^{u_i}}}\right]^\top\]
with $\mathbf{u} := [u_1,u_2,\ldots,u_K]^\top$, and the inner $\sigma$ in \eqref{neural_network} models a general differentiable nonlinearity such as tanh. Although differentiability of the nonlinearities is needed for the derivations in this section, the general idea will be later tested on networks with non-differentiable nonlinearities (such as ReLU) in the experiments. 

Given trained weights $\{\mathbf{W}_l^{(TR)}\}_{l=1}^2 $, and using \eqref{W} and \eqref{M2}, the random weight matrices are found as  
\begin{equation} \label{W_l}
\mathbf{W}_l :=\mathbf{M}_l \text{diag}( \mathbf{z}_{l} ) =  \mathbf{W}_l^{(TR)} \;  \mathbf{S}_l \; \mathbf{D}_l\;\;\;\;l=1,2
\end{equation}  
where $\mathbf{S}_l = \text{diag}( \mathbf{z}_{l} )$ denotes the random sampling matrix with pseudo-inverse diagonal mean given by 
$\mathbf{D}_l = \text{diag}^{\dagger}\Big(\mathbb{E}_{q_z(\mathbf{z}_l; \mathbf{p}_l)} [\mathbf{z}_{l}] \Big)$. Since $\mathbb{ E}[\mathbf{ W}_l ] \mathbf{x}_{(l-1)} = \mathbf{ W}_l^{(TR)} \mathbf{x}_{(l-1)}$, the mean of $\mathbf{W}_l$ does not depend on $\mathbf{p}_l$, while its higher-order moments do. 

\noindent
{\bf Proposition 1.} \emph{For the two-layer network in \eqref{neural_network}, the proposed minimization in \eqref{opt_var} can be  approximated by}
\begin{align}\label{opt_majorized}
\min_{ \{\mathbf{p}_i\geq {\bf 0} , \mathbf{1}^\top \mathbf{p}_i = 1\}_{ i=1}^2} & 	\text{Tr} (\text{Cov}_{\mathbf{W}_2}[\mathbf{W}_2 \sigma(\mathbf{ W}_1^{(TR)} {\mathbf{x}}_\nu) ] )   \\
& + \gamma \text{Tr}( \mathbb{E}_{\mathbf{W}_2}[ \mathbf{W}_2    \mathbf{W}^\top_2]  ) \text{Tr}(\text{ Cov}_{\mathbf{W}_1}[ \mathbf{ W}_1{\mathbf{x}}_\nu ]  ) \nonumber 
\end{align}
\emph{where 
$\gamma$ is a constant. The solution of \eqref{opt_majorized} proceeds in two steps}
\begin{equation*}
\hspace{-1.6cm}{\textbf {Step 1:}} \qquad \quad  \quad \mathbf{p}^*_1 = \arg \min_{ \mathbf{p}_1}  \text{Tr}(\text{ Cov}_{\mathbf{W}_1}[{\mathbf{ W}_1\mathbf{x}}_\nu ]  ) 
\end{equation*}
\begin{eqnarray*}
\hspace{-0.01cm}{\textbf {Step 2:}} \qquad \qquad
\mathbf{p}^*_2 = \arg\min_{ \mathbf{p}_2}  \text{Tr} (\text{Cov}_{\mathbf{W}_2}[\mathbf{W}_2 \sigma(\mathbf{ W}_1^{(TR)} {\mathbf{x}}_\nu ) ] ) \\  + \gamma' \text{Tr}( \mathbb{E}_{\mathbf{W}_2}[ \mathbf{W}_2    \mathbf{W}^\top_2]  ) 
\end{eqnarray*}
where $\gamma' := \gamma\text{Tr}(\text{ Cov}_{\mathbf{W}_1}[{ \mathbf{W}_1 \mathbf{x}}_\nu ]  )\Big|_{\mathbf{p}_1 = \mathbf{p}^*_1 }$.

\noindent
{\emph{ Proof.} See Appendix \ref{sec:app_proof}.}

\noindent
{\bf Remark.} The cost in \eqref{opt_majorized} approximates that in \eqref{opt_var} by casting the overall uncertainty minimization as a weighted sum of layer-wise variances. In particular, $\mathbf{p}_1^*$ is the sampling probability vector  that minimizes variance score of the first layer. It subsequently influences the regularization scalar $\gamma'$ in minimizing the second layer variance, which yields the pmf vector $\mathbf{p}_2^*$. This can be inductively generalized to $L>2$ layers. As $L$ increases however, so do the number of cross terms. For simplicity and scalability, we will further approximate the per-layer minimization by dropping the regularization term, which leads to \emph{separable} optimization across layers. This is an intuitively pleasing relaxation, because layer-wise variance is minimized under $\mathcal{H}_0$, which also minimizes the regularization weight $\gamma'$. 

The resulting non-regularized approximant of step 2 is 
\begin{equation*}\label{sub2_apprx}
\mathbf{p}^*_2 = \arg\min_{ \mathbf{p}_2} 	\text{Tr} (\text{Cov}_{\mathbf{W}_2}[\mathbf{W}_2 \sigma(\mathbf{ W}_1^{(TR)} {\mathbf{x}}_\nu) ] )
\end{equation*}
generalizing to the $l$-th layer in an $L$-layer DNN as 
\begin{equation}\label{layer_l}
\mathbf{p}^*_l = \arg \min_{ \mathbf{p}_l}  \text{Tr}(\text{ Cov}_{\mathbf{W}_l}[{\mathbf{W}_l \mathbf{x}}_{(l-1)} ]  ) 
\end{equation}
where $\mathbf{x}_{(l-1)}$ is the output of the $(l-1)$st layer, regardless of pmf vectors of other layers $\{\mathbf{ p}_{l'}\}_{l'\neq l}$. 

\subsection{Layer-wise variance minimization } 
Here we will solve the layer-wise variance minimization in \eqref{layer_l}. Using \eqref{W_l}, the cost can be upper bounded by 
\begin{align}
\text{Tr}&(\text{ Cov}_{\mathbf{W}_l}[{\mathbf{W}_l \mathbf{x}}_{(l-1)} ]  ) \nonumber \\
&= \mathbb{E}\Big[\|\mathbf{W}_l^{(TR)} \;  \mathbf{S}_l \; \mathbf{D}_l \mathbf{x}_{(l-1)} - \mathbb{E}[\mathbf{W}_l^{(TR)} \;  \mathbf{S}_l \; \mathbf{D}_l \mathbf{x}_{(l-1)}  ]\|_2^2\Big] \nonumber \\
& = \mathbb{E}\Big[\|\mathbf{W}_l^{(TR)} \;  \mathbf{S}_l \; \mathbf{D}_l \mathbf{x}_{(l-1)} - \mathbf{W}_l^{(TR)} \;   \mathbf{x}_{(l-1)} \|_2^2\Big]  \nonumber\\
& \leq \|\mathbf{W}_l^{(TR)}\|_2^2  \; \mathbb{E}\Big[\|  \mathbf{S}_l \; \mathbf{D}_l \mathbf{x}_{(l-1)} -   \mathbf{x}_{(l-1)} \|_2^2\Big]  \nonumber \\
& = \|\mathbf{W}_l^{(TR)}\|_2^2  \; \sum_{i=1}^{h_l} \mathbb{E}[({S}_{ii} {D}_{ii} \mathbf{x}_{(l-1),i} -\mathbf{x}_{(l-1),i})^2]   \nonumber  \\ 
& = \|\mathbf{W}_l^{(TR)}\|_2^2  \;   \sum_{i=1}^{h_l}  \mathbf{x}^2_{(l-1), i}   \mathbb{E}[({S}_{ii} {D}_{ii} -1)^2 ]   \nonumber \\ 
& = \|\mathbf{W}_l^{(TR)}\|_2^2  \; 
\sum_{i=1}^{h_l} \mathbf{x}^2_{(l-1), i} \left(\dfrac{1}{\pi_{l,i}} -1\right)\label{open-close}
\end{align} 
where the last equality follows because the $C$ draws are iid with replacement, and the binary random variables $z_{l,i}$ reduce to Bernoulli ones with parameter $\pi_{l,i} = 1-(1-p_{l,i})^C$; hence, for $\mathbf{x}_{(l-1),i} \neq 0$  it holds that $\mathbb{E}[{S}_{ii}^2 {D}_{ii}^2] = 1/\pi_{l,i}$ and $\mathbb{E}[{S}_{ii} {D}_{ii}]=1$, which implies that $\mathbb{E}[({S}_{ii} {D}_{ii} -1)^2 ]= (1/\pi_{l,i}) -2 +1$.

Using \eqref{open-close}, the optimization in \eqref{layer_l} can be approximately solved by a majorized surrogate as 
\begin{equation}\label{layer_l2}
\min _{\mathbf{p}\geq {\bf 0} , \mathbf{1}^\top \mathbf{p} = 1}   \sum_{i=1}^{h_l}  \dfrac{1}{1-(1-p_{l,i})^C} \mathbf{x}^2_{(l-1), i}
\end{equation}  
which is a convex problem that can be solved efficiently as elaborated next. 

\section{Solving layer-by-layer minimization}\label{sec:solvers}
Consider rewriting the layer-wise variance minimization in \eqref{layer_l2} in a general form as
\begin{equation}\label{opt_before_apprx}
\min _{\mathbf{p}\geq {\bf 0} , \mathbf{1}^\top \mathbf{p} = 1}   \sum_{i=1}^h  \dfrac{\alpha_i}{1-(1-p_i)^{C}} \;.
\end{equation} 
where $\alpha_i:=x_{(l-1), i}^2$ for the $l$-th layer. 
Over the feasible set of the probability simplex, the cost in \eqref{opt_before_apprx} has semi-definite  Hessian; thus, it is convex, and  can be solved by projected gradient descent iterations.  However,  $\mathbf{p}$ lies in the probability simplex space of dimension $h$, the number of hidden nodes in a given layer, and is typically very large. The large number of variables together with possible  ill-conditioning can slow down the convergence rate. 

To obtain a solver with quadratic convergence rate, we build on the fact that $h_l $ is usually very large, which implies that $p_i \ll 1$ for the practical setting at hand. Using the inequality $1-(1-p_i)^C \geq 1-e^{-Cp_i}$, the cost in \eqref{opt_before_apprx} can then be tightly upperbounded, which leads to majorizing \eqref{opt_before_apprx}  as
\begin{equation}\label{opt_general}
\min _{\mathbf{p}\geq {\bf 0} , \mathbf{1}^\top \mathbf{p} = 1}   \sum_{i=1}^h  \dfrac{\alpha_i}{1-e^{-C p_i}} \;.
\end{equation} 
The KKT conditions yield the optimal solution of the convex problem in \eqref{opt_general}, as summarized next. 

\noindent 
{\bf Proposition 2.} \emph{The optimization in \eqref{opt_general} can be solved with quadratic convergence rate, and the optimum is given by }
\begin{equation}\label{p_exact}
p_i^* = -\dfrac{1}{C} \ln\Big(\dfrac{2 \rho^* + x_{(l-1), i}^2  - \sqrt{[2\rho^*+ x_{(l-1), i}^2 ]^2-4\rho^{*2}} }{2\rho^*}\Big)
\end{equation}
\emph{where $\rho^*$ is the solution to the following root-finding problem}
\begin{equation*}
\hspace*{-0.2cm}
\sum_{i=1}^h \ln (2 \rho + x_{(l-1), i}^2 - \sqrt{[2\rho+x_{(l-1), i}^2]^2-4\rho^2}) -n \ln (2\rho) +C = 0 .
\end{equation*}
\emph{Proof.} See Appendix \ref{sec:app_proof2}.

\subsection{Approximate variance minimization for small $C$}
For small values of $C$, it holds that $(1-p_i)^C > 1-Cp_i$; hence, the Bernoulli parameter $\pi_i=1-(1-p_i)^C$ can be approximated by its upperbound $Cp_i > \pi_i$. With this we can approximate the cost in \eqref{layer_l2}, as  
\begin{equation}\label{opt_small_C}
\min _{\mathbf{p}\geq {\bf 0} , \mathbf{1}^\top \mathbf{p} = 1}   \sum_{i=1}^h  \dfrac{\alpha_i}{Cp_i} \;.
\end{equation}  
Using the Lagrangian and the KKT conditions, we then find
$p_i^* = {\sqrt{\alpha_i}}/{\sum_j {\sqrt{\alpha_j}}}$,
which for the $l$-th layer is expressible as 
\begin{equation}\label{p_lin}
p_{(l-1), i}^* = 
\dfrac{|{x_{(l-1),i}}|}{\sum_{j=1}^{h_l} |{x_{(l-1),j}}|}\;.
\end{equation}

This approximation provides analytical justification for the heuristic approach in \cite{dhillon2018stochastic}, where it is proposed to sample with probabilities proportional to the magnitude of the hidden unit outputs. However, there remains a subtle difference, which will be clarified in Section \ref{sec:tests}.

Approximating \eqref{opt_general} with \eqref{opt_small_C} can be loose for large values of $C$, which motivates our next approximation. 

\subsection{Approximate variance minimization for large $C$}
Building on the tight approximation in \eqref{opt_general}, one can further approximate the variance for large $C$ as
\begin{equation}\label{apr2}
\min _{\mathbf{p}\geq {\bf {0}} , \mathbf{1}^\top \mathbf{p} = 1}   \sum_{i=1}^h  \dfrac{\alpha_i}{1-e^{-C p_i}} \simeq \min _{\mathbf{p}\geq \bf 0 , \mathbf{1}^\top \mathbf{p} = 1}   \sum_{i=1}^h  {\alpha_i}{(1+e^{-C p_i})} \nonumber
\end{equation} 
where we have used $(1-\delta)^{-1} \simeq  1+\delta$ as a tight approximation for $0<\delta \ll 1$. 
This leads to the minimization
\[ 
\min _{\mathbf{p}\geq {\bf 0} , \mathbf{1}^\top \mathbf{p} = 1}   \sum_{i=1}^n  {\alpha_i} e^{-C p_i}
\] 
which again is a convex problem, whose solution can be obtained using the KKT conditions that lead to
\[-C \alpha_i e^{-C\hat{p}_i^*} + \lambda =0 \quad \forall i \]
where $\lambda $ is the Lagrange multiplier. Under the simplex constraint on the $\{p_i\}$, this leads to the optimal
\begin{equation}\label{p_log}
\hat{p}^*_i = \Big[\dfrac{1}{C} \ln{[C x_{(l-1), i}^2]} +\hat{\beta}^* \Big]_+
\end{equation}
with $[\,.\,]_+$ denoting the projection on the positive orthant, and the normalization constant $\beta := -\ln\lambda/C$ having optimal value
{
	\[\hat{\beta}^*= \dfrac{C- \sum_{i=1}^h \ln(Cx_{(l-1),i}^2) 1_{\{\hat{p}^*_i>0\}} }{C \sum_{i=1}^h 1_{\{\hat{p}^*_i>0\}} }\;.\]}

Although the solution to the fixed point condition cannot be obtained at one shot, and may require a few iterations to converge, in practice we only perform it once and settle with the obtained approximate solution$\{\hat{p}^*_i\}_{i=1}^h$.

\section{Practical issues}\label{sec:implementation}
The present section deals with efficient implementation of the proposed approach in practice, and establishes links with state-of-the-art { randomization-based} detection methods. 

\subsection{Efficient implementation via non-uniform dropout}\label{sec:nonuniform}
The proposed defense builds on modeling the variational pdf $q_\theta (\omega)$ using a  sampling-with-replacement process. Performing the proposed process however,  may incur overhead complexity during inference when compared to the inexpensive dropout alternative outlined in Sec. \ref{sec:vi}. To reduce this complexity, one can implement our approach using efficient approximations, while leveraging the sampling probabilities learned through our uncertainty minimization. 

Reflecting on the binary variables $\{z_{l,i}\}$ that model the pickup of the hidden node $i$ in the overall sampling process in \eqref{z_l}, one can approximate the joint pmf of $\{z_{l,i}\}_{i=1}^{h_l}$ as
\begin{equation}\label{apprx}
q_z({\bf z}_{l}; {\bf p}_{l}) \simeq \prod_{i=1}^{h_l} q_z(z_{l,i};p_{l,i})
\end{equation}
where random variables $\{z_{l, i }\}_i$ are now viewed as approximately  independent \emph{non-identical} Bernoulli variables with parameters $\{\pi_{l,i}\}_{i=1}^{h_l}$; that is, $z_{l, i } \sim$  Bernoulli$(\pi_{l,i})$ for $i=1,\ldots, h_l$, where $\pi_{l,i} = 1-(1-p_{l,i})^C$. 

Although \eqref{apprx} is an approximation, it provides insight but also an efficient implementation of the sampling process. In fact, the proposed optimization in \eqref{opt_before_apprx} can now be viewed as an optimization over the non-uniform dropout probabilities, coupled implicitly through the hyper-parameter $C$, whose selection guarantees a certain level of randomness. This is to be contrasted with finding optimal dropout probabilities - a task requiring grid search over an $h_l$-dimensional space for layer $l$, where $h_l$ can be hundreds of thousands to millions in CNNs classifying high-quality images. Interestingly, the proposed convex optimization simplifies the high-dimensional grid-search into a scalar root-finding task, whose solution can be efficiently found with super-linear (quadratic) convergence rate. 

\subsection{Placement and adjustment of the sampling units} 
It has been argued that CNN layers at different depths can provide extracted features with variable levels of expressiveness~\cite{garcia2018behavior}. On a par with this, one can envision the defense potential at different depths by incorporating  sampling units across say $B$ blocks of the network as listed in Tables \ref{tab:resnet20} and \ref{tab:resnet34}.   In particular, the dropout defense has been mostly utilized at the last layer after flattening \cite{smith2018understanding}, whereas here we consider the potential of sampling at earlier layers that has gone mostly under-explored so far. This can in turn result in DNN-based classifiers with robustness to adversarial attacks,  as optimal sampling at the initial layers maybe crucial for correct detection of the adversarial input. We henceforth refer to a DNN (or CNN)  equipped  with random sampling as the \emph{detection network}, and the original one without the sampling units as the \emph{full network}.

Similar to the pick up probability $\pi$ in uniform dropouts, the number of draws $C$ in our approach is a hyper parameter that controls the level of randomness present in the detection network. Qualitatively speaking, the smaller number of units (smaller $C$) is picked per layer, the larger `amount of randomness' emerges (further $\pi_{l,i}$ is from $1$). This can lead to forward propagating not as informative (under-sampled) features, meaning not representative of the clean image, and can thus cause unreliable detection. A large $C$ on the other hand, increases the probability to pick up units per layer, which requires a large number of MC realizations for reliable detection, otherwise small randomness will lead to miss-detection. At the extreme, very large $C$ renders the detection and full networks identical, thus leading to unsuccessful detection of adversarial inputs. In a nutshell, there is a trade-off in selecting $C$, potentially different for the initial, middle, and final layers of a given CNN. 

\begin{figure*}
	\centering
	\begin{forest}
		[{Randomization-based} approaches to detecting adversaries in DNNs
		[Uniform dropout\\\scriptsize(Dropout),align=center]
		[Minimum uncertainty (variance) based
		\\\scriptsize(non-uniform sampling),align=center 
		[Fixed/Deterministic
		[Exact\\\scriptsize(VM-exact), align=center , tier=bottom]
		[Linear apprx.\\\scriptsize(VM-lin), align=center  , tier=bottom]
		[Logarithmic apprx.\\\scriptsize(VM-log), align=center , tier=bottom]
		]
		[Dynamic
		[Linear apprx.\\\scriptsize(SAP or DVM-lin) , align=center , tier=bottom]
		[Logarithmic apprx.\\\scriptsize(DVM-log), align=center , tier=bottom]
		]
		]	
		]
	\end{forest}
\vspace{-0.3cm}
	\caption{Overview of { randomization-based} adversary detection schemes}\label{tree}
\end{figure*}
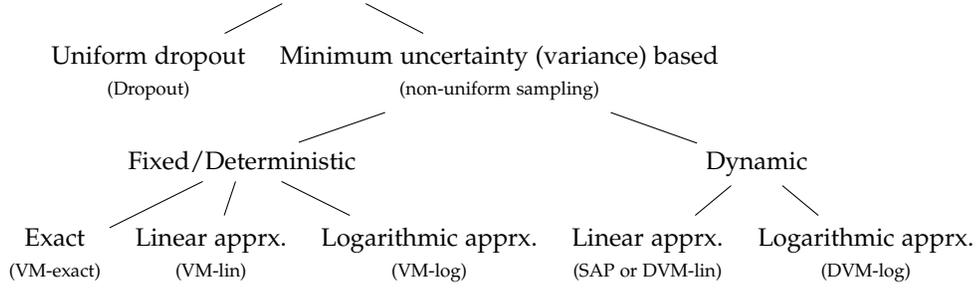

Fig. \ref{tree} categorizes existing and the novel randomization-based approaches to detecting adversarial inputs.  

\noindent
{\bf Uniform dropout.} In this method, units are independently dropped w.p. $1-\pi$, and sampled (picked) w.p. $\pi$ $\forall l,i$. 

\noindent
{\bf Non-uniform dropout using variance minimization.} Dropout here follows the scheme in subsection \ref{sec:nonuniform}, for which we pursue the following two general cases with \emph{deterministic} and \emph{dynamic} probabilities.

\begin{figure}
	\begin{minipage}[b]{0.95\linewidth}
		\centering
		\includegraphics[scale=0.58]{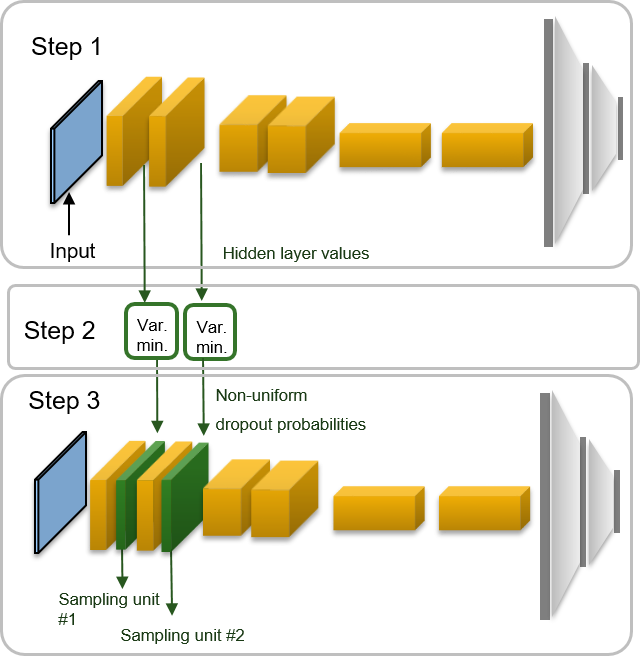}
		
		\centering{	\vspace{0.1cm}	 a) Detection via deterministic sampling probabilities}
	\end{minipage}	
	
	\begin{minipage}[b]{0.95\linewidth}
		\centering
		\vspace{0.2cm}
		\includegraphics[scale=0.5]{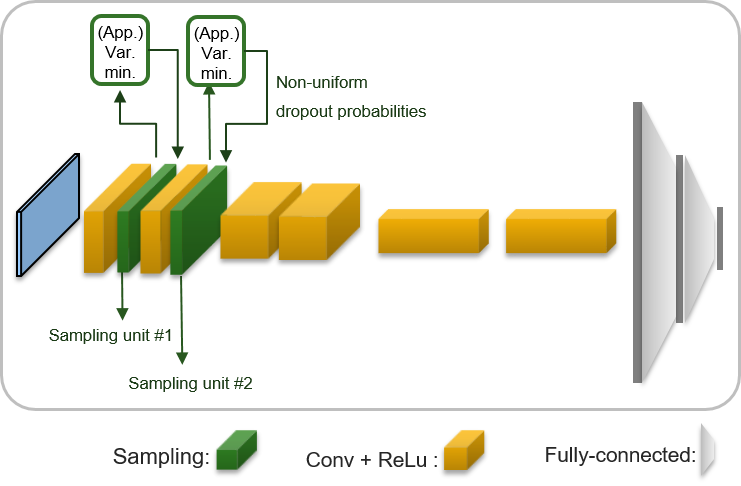}
		
		\centering{	\vspace{0.1cm}	 b) Detection via dynamic sampling probabilities}
	\end{minipage}	
	\caption{Schematic of the proposed detection schemes}\label{fig:schematic}
\end{figure}

{\emph{(C1) Variance minimization with fixed probabilities.}} In this case, the image is first passed through the full network to obtain the values $\{x_{(l-1), i}\}$ of the unit outputs per hidden layer. These are needed to determine the non-uniform dropout probabilities  $1-p_{l, i}$ (thus $\pi_{l,i}$ and then the index of the units to sample) via \emph{exact}, \emph{linear}, or \emph{logarithmic} approximations given in \eqref{p_exact}, \eqref{p_lin} and \eqref{p_log}, respectively, refered to as VM-exact, VM-lin, and VM-log; see Fig. \ref{fig:schematic}-a.

Despite parallel MC passes in the proposed class of sampling with fixed probabilities (step 3 in Fig. \ref{fig:schematic}-a), the first step still imposes a \emph{serial} overhead in  detection since the wanted probabilities must be obtained using a pass through the full network. Our approach to circumventing this overhead is through approximation using the following class of sampling with \emph{dynamic} probabilities. 

{\emph{(C2) Variance minimization with dynamic probabilities.}} Rather than finding the sampling probabilities beforehand, $p_{l,i}^{(r)}$ are determined \emph{on-the-fly} as the image is passed through the detection network with the units sampled per layer. As a result, the observed unit values are random (after passing through at least one unit sampled), and are different across realizations. In order to mitigate solving many optimization problems, variance minimization with dynamic probabilities is only implemented via \emph{linear} and \emph{logarithmic} approximations \eqref{p_lin} and \eqref{p_log}; and are referred to as DVM-lin and DVM-log, respectively; see Fig. \ref{fig:schematic}-b.

\begin{algorithm}[t]\label{alg:deterministic}
	
\SetKwInOut{Input}{Input} 
\caption{Adversary detection - fixed $\{p_{l,i}\}$}
		\Input{Test image  $\mathbf{x}_\nu$, $B,C$, $R$ and $\tau_0$} 
	Pass image $\mathbf{x}_\nu$ through full network; find $\{x_{(l-1), i}\}$\\
	Use $\{x_{(l-1), i}\}$ to obtain $\{p_{l, i}\}$ via  \eqref{p_exact}, \eqref{p_lin} or \eqref{p_log}\\
	\For {$r=1, 2, \ldots,R$}
	{Collect output class $\mathbf{{y}}_\nu^{(r)}$}  
	Estimate the mutual information (MI) of $\{{\mathbf{y}}_\nu^{(r)}\}_{r=1}^R$ \\
	\SetKwInOut{Output}{Output}
	\Output{Declare \emph{adversary} if MI exceeds threshold $\tau_0$}
\end{algorithm}

\begin{algorithm}[]\label{alg:dynamic}
	
	\SetKwInOut{Input}{Input} 
	\caption{Adversary detection - dynamic $\{p_{l,i}\}$}
	\Input{Test image  $\mathbf{x}_\nu$, $B,C$, $R$ and $\tau_0$} 
	\For {$r=1, 2, \ldots,R$}
	{Collect $\mathbf{{y}}_\nu^{(r)}$ after passing $\mathbf{x}_\nu$ through the detection network with units picked with dynamic probabilities obtained (exactly or approximately) 
		using the observed values}  
	Estimate the mutual information (MI) of $\{{\mathbf{y}}_\nu^{(r)}\}_{r=1}^R$ \\
	\SetKwInOut{Output}{Output}
	\Output{Declare \emph{adversary} if MI exceeds threshold $\tau_0$}
\end{algorithm}

It is interesting to note that DVM-lin corresponds to the proposed stochastic activation pruning (SAP) in \cite{dhillon2018stochastic}, with 
\begin{align*}
\mathbf{p}_{l,r}^{\text{SAP}} = \left[\dfrac{|x_{(l-1), 1}^{(r)}|}{\sum_{i=1}^{h_l}|x_{(l-1), i}^{(r)}|}, \dfrac{|x_{(l-1), 2}^{(r)}|}{\sum_{i=1}^{h_l}|x_{(l-1), i}^{(r)}|} , ..., 
\dfrac{|x_{(l-1), h_l}^{(r)}|}{\sum_{i=1}^{h_l}|x_{(l-1), i}^{(r)}|} \right]
\end{align*}
where $x_{(l-1), i}^{(r)}$ is the output of the $i$-th activation unit of the $l$-th layer in the $r$-th realization for input $\mathbf{x}$.

Figure \ref{tree} provides an overview of the sampling methods, while Algorithms \ref{alg:deterministic} and \ref{alg:dynamic} outline the two proposed variance minimization-based detection methods in pseudocode.

\section{Numerical tests}\label{sec:tests}
\begin{algorithm}[]
	\SetKwInOut{Input}{Input}
	\SetKwInOut{Output}{Output}
	\underline{Solve:} $\min_{\mathbf{p}\geq {\bf 0} , \mathbf{1}^\top \mathbf{p} = 1 }   \sum_{i=1}^{h } \dfrac{\alpha_i}{1-(1-p_i)^C} $ 
	\\
	\Input{$[\alpha_1, \alpha_2, \ldots, \alpha_{h}], C$}
	\Output{Nonuniform dropout pmf $\boldsymbol{\pi} = [\pi_1 \ldots \pi_{h}]^\top$}
	
	Using bisection and initialization $\rho_0= \sum_{i=1}^h \alpha_i/h$, find the root $\rho^*$ for $\sum_i \ln (2 \rho + \alpha_i  - \sqrt{(2\rho+\alpha_i)^2-4\rho^2}) -n \ln (2\rho) +C = 0 $  \\
	Set $	p_i^* = -\dfrac{1}{C} \ln\Big(\dfrac{2 \rho^* + \alpha_i - \sqrt{(2\rho^*+\alpha_i)^2-4\rho^{*2}} }{2\rho^*}\Big) \quad \forall i$\\
	Set $\pi_i^* = 1-(1-p_i^*)^C\quad \forall i $

	{\caption{ Layer-wise minimum variance solver}}
	
\end{algorithm}

In this section, we test the effectiveness of the proposed sampling method for detecting various adversarial attacks on CNNs used for image classification. In order to address the raised issue in \cite{carlini2017adversarial}, classification of the CIFAR10 image dataset using ResNet20 as well as the high-resolution cats-and-dogs images using ResNet34 networks \cite{he2016deep} are tested. A short summary of the two networks and datasets can be found in Tables \ref{tab:datasets}, \ref{tab:resnet20} and \ref{tab:resnet34}. In order to investigate the issue around placement of the sampling units,  we will place them after ReLU activation layers in different ``blocks'' ($B$) of the ResNet20 and ResNet34 networks, as listed in Tables \ref{tab:resnet20} and \ref{tab:resnet34}. Numerical tests are made available 
online.\footnote{https://github.com/FatemehSheikholeslami/variance-minimization}

\begin{table}[]
	\begin{center}\begin{tabular}{ |c|c|c|c| c| } 
			\hline
			Dataset & image size & \# train &\# val. & \# test\\ 
			\hline		
			CIFAR10 & 32 x 32  & 50,000 & 2,000 & 8,000\\ 
			\hline	
			Cats-and-dogs &  224 x 224 & 10,000 &2,000 & 13,000 \\ 
			\hline		
		\end{tabular}
		\caption{CIFAR10 and cats-and-dogs image-classification datasets}
			\label{tab:datasets}
	\end{center}
\end{table}

\begin{table}
	\begin{center}\begin{tabular}{ |c|c|c|c|  }
			\hline
			name & output-size & 20 layers &\multirow{2}{4.2em}{\centering  \#sampling units }\\ 
			& & & \\
			\hline
			Block1 & 32 x 32 & [ 3 x 3, 16] & 1  \\
			\hline
			Block2 & 32 x 32 & $ \begin{bmatrix}
			3 \times 3, 16 \\
			3 \times 3, 16
			\end{bmatrix} \times 3$& 6 \\
			\hline
			Block3 &  16 x 16& $ \begin{bmatrix}
			3 \times 3, 32 \\
			3 \times 3, 32
			\end{bmatrix} \times 3$& 6  \\
			\hline
			Block4 & 8 x 8  & $ \begin{bmatrix}
			3 \times 3, 64 \\
			3 \times 3, 64
			\end{bmatrix} \times 3$ & 6 \\
			
			\hline
			
			 &  &   \multirow{3}{7.4em}{average pool, \\$64$-d fully conn.,\\ softmax}& \\
			Block5& 1 x 1 & &  1\\
			& & & \\
			\hline
		\end{tabular}
		\caption{ResNet20 architecture on CIFAR10 dataset}
		 \label{tab:resnet20}	
	\end{center}
\end{table}

\begin{table}[] 
	\begin{center}\begin{tabular}{ |c|c|c|c|  } 
			\hline
			& output-size & 34 layers & \multirow{2}{4.2em}{\centering  \#sampling units }\\ 
			& & & \\
			\hline
			Block 1 & 112 x 112 &  \multirow{2}{7.2em}{ [7 x 7, 64],\\ 3x3 max-pool} & 2\\
			& &  & \\
			\hline
			Block2 & 56 x 56 & $ \begin{bmatrix}
			3 \times 3, 64 \\
			3 \times 3, 64
			\end{bmatrix} \times 3$& 6\\
			\hline
			Block3 & 28 x 28 & $ \begin{bmatrix}
			3 \times 3, 128 \\
			3 \times 3, 128
			\end{bmatrix} \times 4$&8 \\
			\hline
			Block4 & 14 x 14 & $ \begin{bmatrix}
			3 \times 3, 256 \\
			3 \times 3, 256
			\end{bmatrix} \times 6$& 12\\
			\hline
			Block5 & 7 x 7 & $ \begin{bmatrix}
			3 \times 3, 512 \\
			3 \times 3, 512
			\end{bmatrix} \times 3$& 6 \\
			\hline
			
			&  &   \multirow{3}{7.2em}{ average pool, \\ $1000$-d fc, softmax} & \\
			Block 6 & 1 x 1 &  &1 \\
			& & & \\
			\hline
		\end{tabular}
		\caption{ResNet34 architecture on cats-and-dogs dataset}
			\label{tab:resnet34}
	\end{center}
\end{table}

\subsection{CIFAR10 dataset}
ResNet20 is trained using $20$ epochs with minibatches of  size $128$. Adversarial inputs 
are crafted on the corresponding MC network as in \cite{smith2018understanding}, using the fast gradient sign method (FGSM) \cite{goodfellow2014explaining}, the basic iterative method (BIM) \cite{kurakin2016adversarial}, the momentum iterative method (MIM) \cite{dong2018boosting}, and the Carlini-and-Wagner (C\&W) \cite{carlini2017towards} attacks. 
Parameters of the attacks as well as test accuracy of the MC network on clean and adversarial inputs are reported in Table \ref{tab:attackscifar}. 

Placement parameter $B$ and sampling parameters $C$ for variance minimization methods as well as the dropout probability for uniform dropout are selected by cross validation. To clarify the suboptimality gap between the exact and approximate variance minimization with {\bfseries{deterministic}} sampling probabilities, we have cross-validated the parameters for VM-exact, and reused  them for VM-lin and VM-log approximates. 

The sampling parameter is selected as $C=f \times \text{nnz}(\mathbf{x}_l)$ for the $l$-th layer sampling unit, where $\text{nnz}(.)$ denotes the number of non-zero entries\footnote{This selection is chosen by taking into account the fact that, only non-zero samples will be dropped upon not being selected, while zero entries will remain unchanged regardless of the sampling outcome.}, and $f$ is the sampling ratio varied in $f\in \{0.6, 0.7, 0.8, 0.9, 1.0, 1.5, 2.0, 3.0, 4.0\}$.\footnote{Since the sampling procedure is modeled with replacement, fraction $f$ may be selected greater than 100\%.} Probability in uniform dropout is also varied as $\pi_{\text{drop}} \in \{0.1, 0.2, \ldots, 0.7\}$, and the number of MC runs is $R=20$. 

\begin{figure*}
	\begin{minipage}[b]{0.35\linewidth}
		\hspace{-0.7cm}
		\includegraphics[scale=0.6]{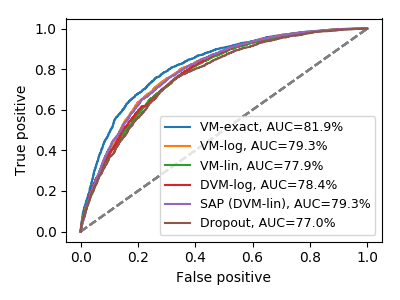}
		
		\centering{	\hspace{-1cm}	 a) FGSM attack with $\epsilon=10$}
	\end{minipage}	
	\begin{minipage}[b]{0.35\linewidth}
		\hspace{-0.8cm}	
		\includegraphics[scale=0.6]{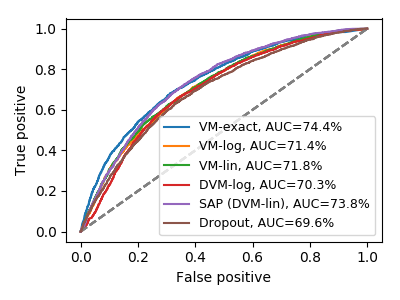}
		
		\centering{	\hspace{-1cm}	 b) MIM attack with $\epsilon=10$}
	\end{minipage}	
	\begin{minipage}[b]{0.35\linewidth}
		\hspace{-1cm}
		\includegraphics[scale=0.6]{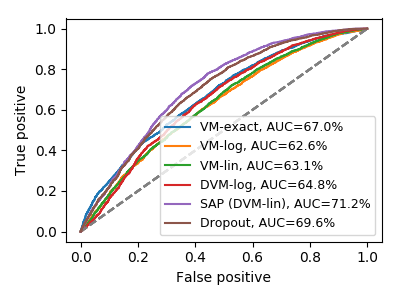}
		
		\centering{	\hspace{-1cm}	 c) BIM attack with $\epsilon=10$}
	\end{minipage}	
	\begin{minipage}[b]{0.35\linewidth}
		\hspace{-0.7cm}		\includegraphics[scale=0.6]{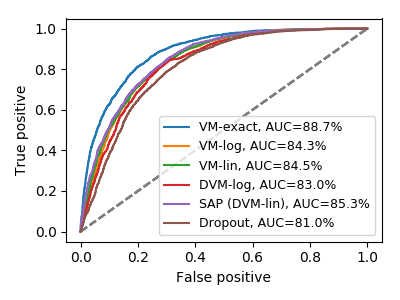}
		
		\centering{\hspace{-1cm} d) FGSM attack with $\epsilon=20$}
	\end{minipage}	
	\begin{minipage}[b]{0.35\linewidth}
		\hspace{-0.8cm}		\includegraphics[scale=0.6]{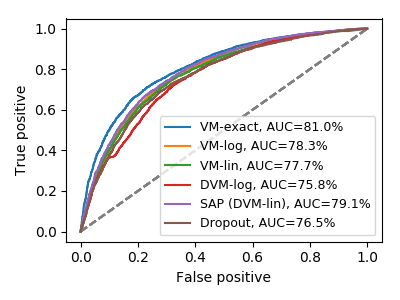}	
		
		\centering{\hspace{-1cm} e) MIM attack with $\epsilon=20$}
	\end{minipage}	
	\begin{minipage}[b]{0.35\linewidth}
		\hspace{-1cm}	\includegraphics[scale=0.6]{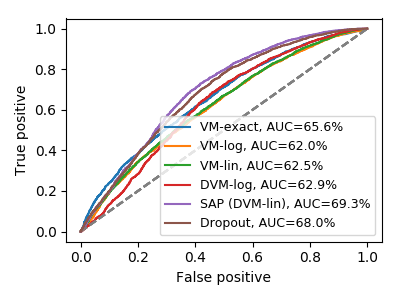}
		
		\centering{\hspace{-1cm} f) BIM attack with $\epsilon=20$}
	\end{minipage}	
	\begin{minipage}[b]{0.4\linewidth}
		\hspace{+2cm}	\includegraphics[scale=0.6]{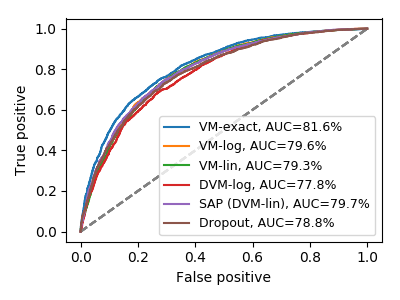}
		
		\centering{\hspace{+4cm} g) C\&W attack}
	\end{minipage}	
	\begin{minipage}[b]{0.4\linewidth}
		\hspace{+1cm}	\includegraphics[scale=0.6]{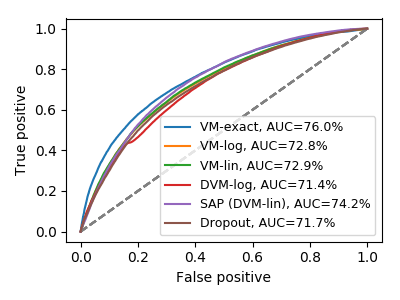}	
		
		\centering{\hspace{+2cm} h) Combination attack}
	\end{minipage}

%

	\caption{ROC-curve of different attack-detection sampling schemes on CIFAR10 dataset against different attacks. }
	\label{fig:roc-cifar}
\end{figure*}

In order to properly evaluate accuracy in detection of adversarial images, we only aim at detecting the test samples that are correctly classified by the full network, and misclassified after the adversarial perturbation. The detection performance is then reported in terms of the receiver operating characteristic (ROC) curve  in Fig. \ref{fig:roc-cifar}, obtained by varying the threshold parameter $\tau_0$. The exact area-under-curve values along with parameters $B,f,\pi_{\text{drop}}$ are also reported in Tables \ref{tab:auc-cifar1} and \ref{tab:auc-cifar2},  highlighting the improved detection via the proposed variance minimization approach.  

Furthermore, in order to target more realistic scenaria, where attack generation is unknown and may indeed be crafted via various methods, we have also tested the performance against a ``combination attack,'' in which the adversarial input crafted with all 7 settings of attacks are considered.  
This indeed corroborates that placement of the sampling units in the fourth block along with careful tuning of the sampling probabilities via VM-exact provides the highest curve against combination of attacks, while its approximations follow in performance, outperforming uniform dropout. 
For further discussion on sensitivity against parameter selection, see Appendix \ref{sec:app_sensitivity}.

\begin{table*}[]
	\begin{center}\begin{tabular}{ |c|c| c|c|c|c| c|c|c| } 
			\hline
			&clean & \multicolumn{2}{c|}{FGSM}   & \multicolumn{2}{c|}{BIM}  & \multicolumn{2}{c|}{MIM}  & C\&W \\ 
			\hline
			&  &   \multicolumn{2}{c|}{}   & \multicolumn{2}{c|}{norm: $\infty$} & \multicolumn{2}{c|}{	norm: $\infty$}&  
			\# binary search: 10
			\\
			Attack &  -- & \multicolumn{2}{c|}{--}   & \multicolumn{2}{c|}{\# iter: 20} & \multicolumn{2}{c|}{	\# iter: 20}& 	\#max iter: 20  \\
			parameters &   & \multicolumn{2}{c|}{}   & \multicolumn{2}{c|}{$\epsilon_{iter}$: 1/255 } & \multicolumn{2}{c|}{{$\epsilon_{iter}$: 1/255 }}& learning rate:0.1 \\
			\cline{3-8}
			& & \small{$\epsilon=\small{10/255}$}& \small$\epsilon=\small{20/255}$ & \small$\epsilon=\small{10/255}$ &\small$\epsilon=\small{20/255}$ & \small$\epsilon=\small{10/255}$ & \small$\epsilon=\small{20/255}$ &  initial const.: 10 	\\
			\hline
			Class.	 Acc.& 91.5\% & 64.87\% & 56.91\%  &  5.2\%&  5.0\%& 5.4\% & 5.1\% &  11.7\% \\	
			\hline		
		\end{tabular}
		\caption{{\color{black}Attack parameters and test accuracy on clean and adversarial input in CIFAR10 dataset.}}
		\label{tab:attackscifar}
	\end{center}
\end{table*}

\begin{table*}[!h]
	\begin{center}
		\small
		\begin{tabular}{ |c||c |c| c|c||c|c|c|c| } 
			\hline
			&\multicolumn{4}{c||}{ FGSM Attack} & \multicolumn{4}{c|}{ MIM Attack}  \\
			
			Sampling & \multicolumn{2}{c|}{ $\epsilon=10$ } & \multicolumn{2}{c||}{ $\epsilon=20$ } & \multicolumn{2}{c|}{ $\epsilon=10$ }& \multicolumn{2}{c|}{ $\epsilon=20$ } \\ 
			\cline{2-9}
			Method & Parameters & AUC  & Parameters  & AUC  & Parameters & AUC  & Parameters & AUC  \\ 
			
			\hline
			
			\footnotesize VM   & \multirow{3}{4em}{}  &  \bf 81.9 & 
			& \bf 88.7& & \bf  74.4
			&  & \bf 81.0  \\ 
			\cline{1-1}  \cline{3-3} \cline{5-5} \cline{7-7} \cline{9-9} 
			\footnotesize VM-log 
			& \small $(B,f)=(4,2.0)$  &  79.3
			&  \small $(B,f)=(4,4.0)$& 84.3
			& \small $(B,f)=(4,4.0)$ & 71.4
			& \small $(B,f)=(4, 4.0)$ &  78.3 \\ 
			\cline{1-1}  \cline{3-3} \cline{5-5} \cline{7-7} \cline{9-9} 
			\footnotesize VM-linear 
			&  & 77.9 & 
			& 84.5 &  & 71.8
			&  & 77.7 \\ 
			\hline
			\footnotesize DVM-log 
			&\small $(B,f)=(4,3.0)$  & 78.4
			& \small $(B,f)=(5,0.7)$ & 83.0
			& \small $(B,f)=(5,4.0)$ & 70.3
			&\small $(B,f)=(4, 4.0)$  & 75.8 \\ 
			
			\hline
			\footnotesize SAP 
			& \small $(B,f)=(2,4.0)$ &  79.3
			&\small $(B,f)= (3,4.0)$	& 85.3
			& \small $(B,f)= (2,3.0)$  & 73.8
			& \small $(B,f)= (3,4.0)$  & 79.1 \\ 
			\hline
			\footnotesize Dropout
			& \small $(B,\pi_{\text{\tiny drp}})=(5,0.1)$ &  77.0
			&\small $(B,\pi_{\text{\tiny drp}})=(5,0.1)$ 	& 81.0
			& \small $(B,\pi_{\text{\tiny drp}})=(5, 0.1)$  & 69.6
			& \small $(B,\pi_{\text{\tiny drp}})=(5,0.2)$  & 76.5 \\ 
			\hline
		\end{tabular}
	\end{center}
	
	\caption{{\color{black} AUC-ROC of different attack-detection sampling schemes on CIFAR10 test set against FGSM and MIM attacks. Higher values indicate better detection. } }
	\label{tab:auc-cifar1}
\end{table*}

\begin{table*}[!h]
\begin{center}
\small
\begin{tabular}{ |c||c |c| c|c||c|c||c|c| } 
\hline
&\multicolumn{4}{c||}{ BIM Attack} & \multicolumn{2}{c||}{ C\&W Attack} & \multicolumn{2}{c|}{Combination  Attack}  \\

Sampling &\multicolumn{2}{c|}{$\epsilon=10$}&\multicolumn{2}{c||}{$\epsilon=20$}& \multicolumn{2}{c||}{}&\multicolumn{2}{c|}{}\\ 
\cline{2-9}

Method& parameters& AUC  & parameters & AUC  & parameters & AUC  &  parameters & AUC  \\ 

\hline

{\footnotesize {VM}}   & \multirow{3}{4em}{}  &  67.0 & 	& 65.6 & & \bf 81.6	&  & \bf  76.0 \\ 
\cline{1-1}  \cline{3-3} \cline{5-5} \cline{7-7} \cline{9-9}

\footnotesize VM-log&\small$(B,f)=(4,4.0)$& 62.6  
& \small$(B,f)=(4,4.0)$& 62.0
&\small$(B,f)=(4, 3.0)$ & 79.6 
& \small$(B,f)=(4, 4.0)$ & 72.8\\ 			
\cline{1-1}  \cline{3-3} \cline{5-5} \cline{7-7} \cline{9-9} 

\footnotesize VM-linear &  & 63.1 & 	& 62.5 &  &	 79.3&  & 72.9 \\ 
\hline

\footnotesize DVM-log & 
\small$(B,f)=(1, 3.0)$ & 64.8 &
 \small$(B,f)=(1, 4.0)$&  62.9 & 
 \small$(B,f)=(5, 0.8)$ & 77.8	& 
 \small$(B,f)=(4,3.0)$ & 71.4 \\ 
\hline

\footnotesize SAP& 
\small$(B,f)=(2,1.5)$ & \bf 71.2 &
\small$(B,f)=(2, 1.5)$ & \bf 69.3 & 
\small$(B,f)=(5,4.0)$ & 79.7	& 
\small$(B,f)=(2,3.0)$ & 74.2 \\ 
\hline
\footnotesize Dropout	& 
\small$(B,\pi_{\text{\tiny drp}})=(2,0.1)$ & 69.6 & \small$(B,\pi_{\text{\tiny drp}})=(2,0.1)$& 68.0 & \small$(B,\pi_{\text{\tiny drp}})=(5,0.2)$ & 78.8 &\small$(B,\pi_{\text{\tiny drp}})=(5,0.1)$  & 71.7 \\ 
\hline
\end{tabular}
\end{center}

\caption{{\color{black}AUC-ROC of different attack-detection sampling schemes on CIFAR10 test set against FGSM, C\&W, and combination attacks. Higher values indicate better detection.} }
\label{tab:auc-cifar2}
\end{table*}

\subsection{Cats-and-dogs dataset}

\begin{table*}[]
	\begin{center}\begin{tabular}{ |c|c| c|c|c|c| c|c|c| } 
			\hline
			&clean & \multicolumn{2}{c|}{FGSM}   & \multicolumn{2}{c|}{BIM}  & \multicolumn{2}{c|}{MIM}  & C\&W \\ 
			\hline
			&  &   \multicolumn{2}{c|}{}   & \multicolumn{2}{c|}{norm: $\infty$} & \multicolumn{2}{c|}{	norm: $\infty$}&  
			\# binary search steps: 10
			\\
			Attack &  -- & \multicolumn{2}{c|}{--}   & \multicolumn{2}{c|}{\# iter: 10} & \multicolumn{2}{c|}{	\# iter: 20}& 	\#max iter: 20  \\
			parameters &   & \multicolumn{2}{c|}{}   & \multicolumn{2}{c|}{$\epsilon_{iter}$: 1.5 } & \multicolumn{2}{c|}{{$\epsilon_{iter}$: 1.5}}& learning rate: 0.1  \\
			\cline{3-8}
			& & \small{$\epsilon={10}$}& \small$\epsilon={20}$ & \small$\epsilon={10}$ &\small$\epsilon={20}$ & \small$\epsilon={10}$ & \small$\epsilon={20}$ &  initial const.: 10 	\\
			\hline
			Class.	 Acc.& $94.5\%$ & $74.85\%$ & $70.5\%$ & $19.2\%$ & $18.4\%$& $12.9\%$&  $9.9\%$& $68.95\%$ \\	
			\hline		
		\end{tabular}
	\caption{	{\color{black} Attack parameters and test accuracy on clean and adversarial input in cats-and-dogs dataset.}}
		\label{tab:attackcad}
	\end{center}
\end{table*}
 
 \begin{table*}[!h]
 	\begin{center}
 		\small
 		\begin{tabular}{ |c||c |c| c|c||c|c|c|c| } 
 			\hline
 			&\multicolumn{4}{c||}{ FGSM Attack} & \multicolumn{4}{c|}{ MIM Attack}  \\
 			
 			Sampling & \multicolumn{2}{c|}{ $\epsilon=10$ } & \multicolumn{2}{c||}{ $\epsilon=20$ } & \multicolumn{2}{c|}{ $\epsilon=10$ }& \multicolumn{2}{c|}{ $\epsilon=20$ } \\ 
 			\cline{2-9}
 			Method & Parameters & AUC  & Parameters  & AUC  & Parameters & AUC  & Parameters & AUC  \\ 
 			
 			\hline
 			
 			\footnotesize VM   & \multirow{3}{4em}{}  & { 73.5 }& 
 			& 84.0 & & \bf 72.1
 			&  &  78.5  \\ 
 			\cline{1-1}  \cline{3-3} \cline{5-5} \cline{7-7} \cline{9-9} 
 			\footnotesize VM-log 
 			& \small $(B,f)=(2,0.7)$  &   57.2
 			&  \small $(B,f)=(2,0.7)$& 63.8
 			& \small $(B,f)=(2,0.7)$ & 58.1
 			& \small $(B,f)=(2,0.7)$ &  61.5 \\ 
 			\cline{1-1}  \cline{3-3} \cline{5-5} \cline{7-7} \cline{9-9} 
 			\footnotesize VM-linear 
 			&  & 70.8 & 
 			& 82.2 &  & 68.8
 			&  & 77.1 \\ 
 			\hline
 			\footnotesize DVM-log 
 			&\small $(B,f)=(4,3.0)$  & \bf 76.1
 			& \small $(B,f)=(4,3.0)$ & \bf 84.5
 			& \small $(B,f)=(6,0.7)$ & 68.5
 			&\small $(B,f)=(1,2.0)$  & \bf 83.9 \\ 
 			
 			\hline
 			\footnotesize SAP 
 			& \small $(B,f)=(6,0.7)$ &  70.3
 			&\small $(B,f)=(6,2.0)$	& 83.7
 			& \small $(B,f)=(1,3.0)$  & 69.4
 			& \small $(B,f)=(6,1.0)$  & 74.8 \\ 
 			\hline
 			\footnotesize Dropout
 			& \small $(B,\pi_{\text{\tiny drp}})=(6,0.3)$ & 70.1 
 			&\small $(B,\pi_{\text{\tiny drp}})=(6,0.1)$	& 83.3
 			& \small $(B,\pi_{\text{\tiny drp}})=(5,0.1)$  & 63.6
 			& \small $(B,\pi_{\text{\tiny drp}})=(6,0.1)$  & 73.7 \\ 
 			\hline
 		\end{tabular}
 	\end{center}
 	
 	\caption{	{\color{black}AUC-ROC of different attack-detection sampling schemes on cats-and-dogs dataset with against FGSM and MIM attacks. Higher values indicate better detection.} } 
 	\label{tab:auc-cad1}
 \end{table*}

\begin{figure*}
	\begin{minipage}[b]{0.35\linewidth}
		\hspace{-0.7cm}
		\includegraphics[scale=0.6]{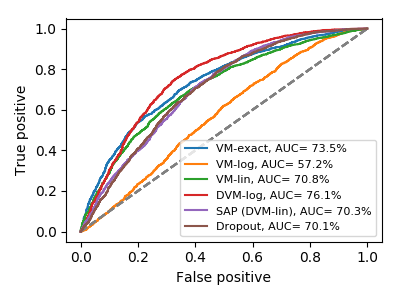}
		
		\centering{	\hspace{-1cm}	 a) FGSM attack with $\epsilon=10$}
	\end{minipage}	
	\begin{minipage}[b]{0.35\linewidth}
		\hspace{-1cm}	
		\includegraphics[scale=0.6]{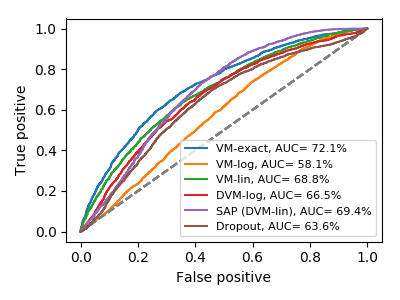}
		
		\centering{	\hspace{-1cm}	 b) MIM attack with $\epsilon=10$}
	\end{minipage}	
	\begin{minipage}[b]{0.35\linewidth}
		\hspace{-1cm}
		\includegraphics[scale=0.6]{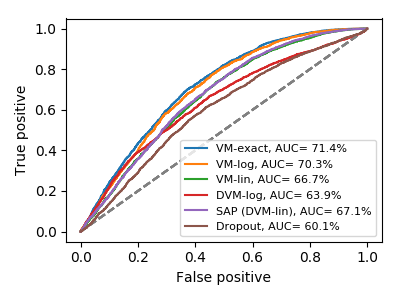}
		
		\centering{	\hspace{-1cm}	 c) BIM attack with $\epsilon=10$}
	\end{minipage}	
	\begin{minipage}[b]{0.35\linewidth}
		\hspace{-0.7cm}		\includegraphics[scale=0.6]{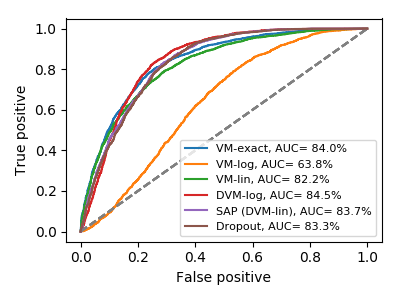}
		
		\centering{\hspace{-1cm} d) FGSM attack with $\epsilon=20$}
	\end{minipage}	
	\begin{minipage}[b]{0.35\linewidth}
		\hspace{-1cm}		\includegraphics[scale=0.6]{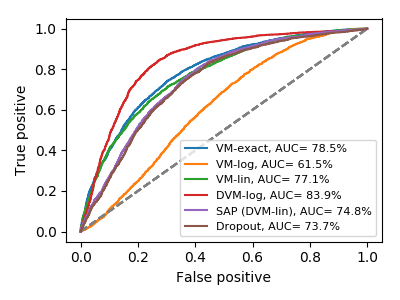}	
		
		\centering{\hspace{-1cm} e) MIM attack with $\epsilon=20$}
	\end{minipage}	
	\begin{minipage}[b]{0.35\linewidth}
		\hspace{-1cm}	\includegraphics[scale=0.6]{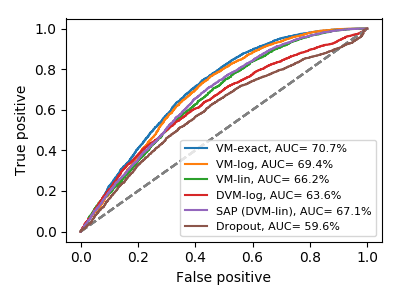}
		
		\centering{\hspace{-1cm} f) BIM attack with $\epsilon=20$}
	\end{minipage}	
	\begin{minipage}[b]{0.4\linewidth}
		\hspace{+2cm}	\includegraphics[scale=0.6]{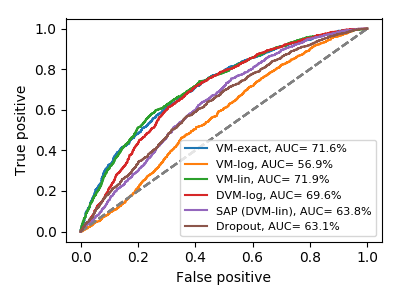}
		
		\centering{\hspace{+4cm} g) C\&W attack}
	\end{minipage}	
	\begin{minipage}[b]{0.4\linewidth}
		\hspace{+1cm}	\includegraphics[scale=0.6]{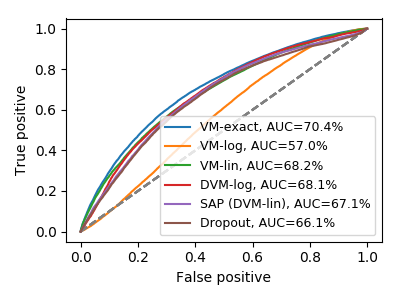}	
		
		\centering{\hspace{+2cm} h) Combination attack}
	\end{minipage}		
	
	\caption{ROC-curve of different attack-detection sampling schemes on cats-and-dogs dataset against different attacks. }
	\label{fig:roc-cad}
\end{figure*}

Tests are also carried out for the cats-and-dogs dataset,\footnote{ https://www.microsoft.com/en-us/download/details.aspx?id=54765} which consists of high-quality images classified into binary classes of cats and dogs. Images are resized to $224 \times 224$, and are classified using ResNet34 \cite{he2016deep}. Weights of the convolutional layers are transferred from the network trained on the ImageNet 
dataset.\footnote{https://github.com/qubvel/classification\_models} This is subsequently followed by a dropout, $1000\times 2$ fully-connected and softmax layer, whose weights are trained using $10,000$ images; see Table \ref{tab:datasets}. 
The FGMS, BIM, MIM, an C\&W attacks are crafted, and parameters are reported in Table \ref{tab:attackcad}. Detection parameters are similarly selected by using the validation set and varying $B \in \{1,2,3,4,5,6\}$, $f\in \{0.6, 0.7, 0.8, 0.9, 1.0, 1.5, 2.0, 3.0\}$, where $C=f \times \text{nnz}(\mathbf{x}_l)$, $\pi_{\text{drop}} \in \{0.1, 0.2, \ldots, 0.7\}$, and the number of MC runs is $R=20$. 

Fig. \ref{fig:roc-cad} plots the ROC curve for detection of adversarial versus clean images, and defense parameters are reported in Tables \ref{tab:auc-cad1} and \ref{tab:auc-cad2}, quantifying the accuracy of attack detection across different methods.  As with CIFAR10, tests are also extended to a combination attack, where detection is performed against the combination of all seven attacks with a fixed set of defense parameters $B,f$, and $\pi_{\text{drop}}$. 

Interestingly, it is observed that for small values of $f$, the linear approximation for variance minimization (VM-lin) follows the performance of the exact variance minimization (VM-exact) closely for FGSM, MIM, and C\&W attacks, whereas the logarithmic approximation (VM-log) exhibits a large gap in performance. In contrast, for large values of $f$, VM-log demonstrates a smaller optimality gap with VM-exact as opposed to VM-lin; see Figs. \ref{fig:roc-cad} (c) and (f). This corroborates our approximations in  \eqref{opt_small_C} and \eqref{p_log}, providing high-performance low-complexity substitutes for the exact variance-minimization solver in both small and large sampling regimes, that is  $f<1$ and $f>1$. Similarly, improved performance of the logarithmic approximates versus the linear ones are also corroborated in the high-quality cats-and-dogs images versus CIFAR10, due to higher $C$  resulting from higher dimensional vectors $\mathbf{x}_{(l)}$ in the hidden layers. 

The ROC curves further demonstrate that performance of the deterministic sampling probabilities, obtained by passing the image through the full network, are often superior  to the dynamic ones (SAP and DVM-log), among which DVM-log demonstrates better performance. 

\subsection{Detection of adaptive attacks}
{In order to further evaluate the performance of the proposed detection schemes against white-box adaptive attacks, we now consider an adaptive attack setting, in which the attacker is aware of the defense mechanism, and designs the adversarial perturbation to jointly fool the classifier and the detector.}

\begin{figure*}

		\begin{minipage}[b]{0.35\linewidth}
			\hspace{-0.7cm}
			\includegraphics[scale=0.6]{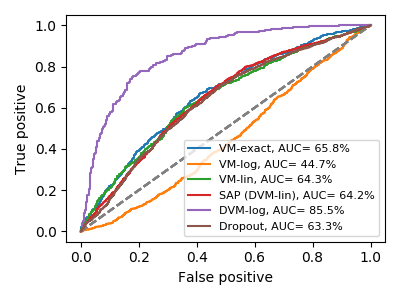}
			
			\centering{	\hspace{-1cm}		 a)  Adaptive-FGSM  $\epsilon=10, \mu=0.01$}
		\end{minipage}	
		\begin{minipage}[b]{0.35\linewidth}
			\hspace{-1cm}	
			\includegraphics[scale=0.6]{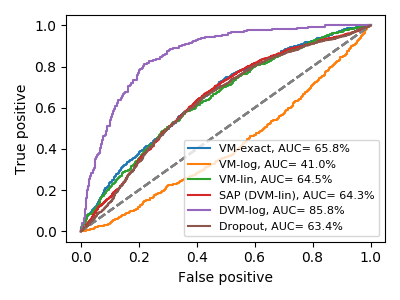}
			
			\centering{	\hspace{-1cm}		 b)  Adaptive-FGSM  $\epsilon=10, \mu=0.1$}
		\end{minipage}	
		\begin{minipage}[b]{0.35\linewidth}
			\hspace{-1cm}
			\includegraphics[scale=0.6]{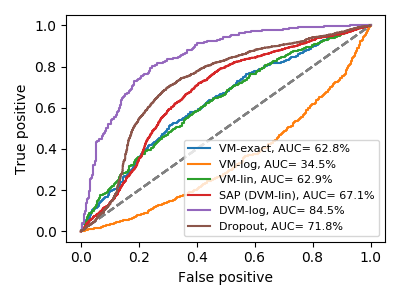}
			
			\centering{	\hspace{-1cm}		c)  Adaptive-FGSM  $\epsilon=10, \mu=1$}
		\end{minipage}	
		\begin{minipage}[b]{0.35\linewidth}
			\hspace{-0.7cm}		\includegraphics[scale=0.6]{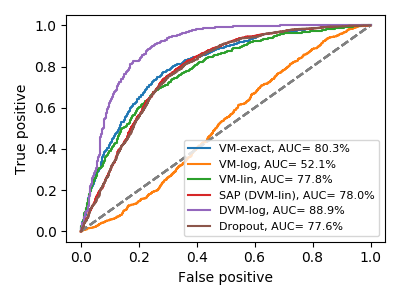}
			
			\centering{\hspace{-1cm} d) Adaptive-FGSM  $\epsilon=20, \mu=0.01$}
		\end{minipage}	
		\begin{minipage}[b]{0.35\linewidth}
			\hspace{-1cm}		\includegraphics[scale=0.6]{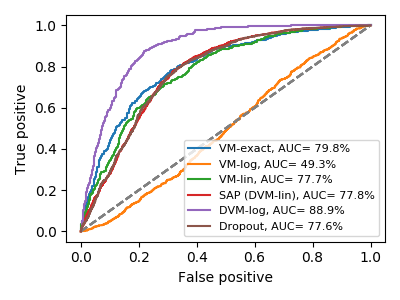}	
			
			\centering{\hspace{-1cm}	 e)  Adaptive-FGSM  $\epsilon=20, \mu=0.1$}
		\end{minipage}	
		\begin{minipage}[b]{0.35\linewidth}
			\hspace{-1cm}	\includegraphics[scale=0.6]{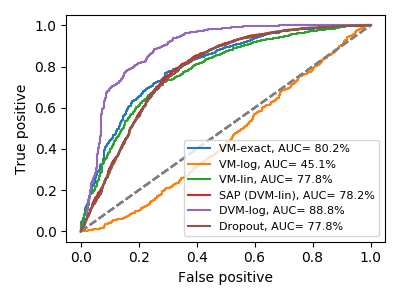}
			
			\centering{\hspace{-1cm}	 f)  Adaptive-FGSM  $\epsilon=20, \mu=1$}
		\end{minipage}	
			\caption{ROC-curve for different values of $\mu$ and $\epsilon$ in adaptive attacks on the cats-and-dogs dataset.}
		\label{fig:roc-adaptive2}
				\vspace{-0.2cm}
	\end{figure*}

\begin{figure*}
		\vspace{-0.2cm}
		\centering
		\includegraphics[scale=0.7]{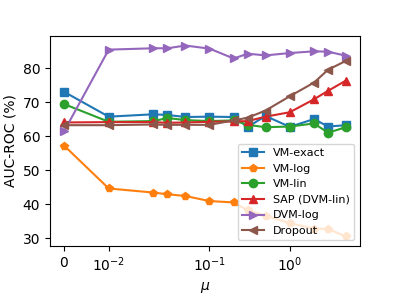}%
		\includegraphics[scale=0.65]{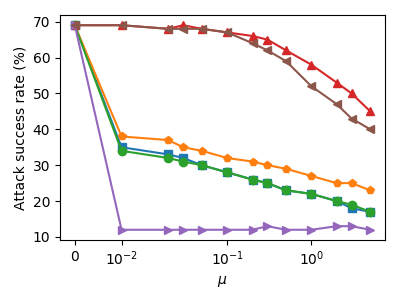}%
				\vspace{-0.5cm}
		\caption{AUC-ROC and attack success rate against adaptive FGSM  attack with $\epsilon=10$ for varying values of $\mu$ for the cats-and-dogs dataset.}
		\label{fig:auc_and_acc2}
	\end{figure*}

{Specifically, let us now model the attacker by seeking perturbation $\delta$ such that not only classification error, namely} {cross-entropy, is maximized, but also uncertainty, or variance, of the network output is simultaneously minimized.

Thus, the adversarial objective is updated as  
		\begin{align}\label{attack_final2}
		\nonumber \min_{\|\delta\|_\infty <\epsilon}  & \quad \sum_{k=1}^K  y_{\nu,k}^{\text{target}} \log(y_{\nu,k}(\mathbf{x}+ \boldsymbol{\delta }))\\ &  \quad + \dfrac{\mu}{R} \sum_{r=1}^R \|\mathbf{y}_{\nu}^{(r)}(\mathbf{x}_\nu+ \boldsymbol{\delta }) - \bar{ \mathbf{y}}_{\nu}(\mathbf{x}_\nu+ \boldsymbol{\delta })  \|_2^2\:.
		\end{align}}
{		The first term in the objective is the negative cross-entropy between the ground-truth one-hot label of input $\mathbf{x}_\nu$, denoted by $\mathbf{y}_{\nu}^{\text{target}} = [y_{\nu,1}^{\text{target}}, \ldots, y_{\nu,K}^{\text{target}}]$, and the soft-max output of the deterministic deep neural network $y_{\nu,k}(\mathbf{x}+ \boldsymbol{\delta })$, that is with no random sampling unit. The second term is the estimated variance of the detection network output over $r=1,...,R$ realizations, denoted by $\mathbf{y}_{\nu}^{(r)}$, whose expected value is estimated by its sample average as
		\begin{equation}
		{\bar{\mathbf{y}}}_{\nu}(\mathbf{x}_\nu+ \boldsymbol{\delta }):= \dfrac{1}{R} \sum_{r=1}^R  \mathbf{y}_{\nu}^{(r)}(\mathbf{x}_\nu+ \boldsymbol{\delta })  
		\end{equation}
		Finally, the scalar $\mu$ balances the trade-off between maximizing the classification error (chance of a successful attack), and minimizing the output uncertainty, thus mitigating detection. This can potentially lead to lower attack success rate as misclassification is not the sole objective anymore.
	
	Solving \eqref{attack_final2} analytically is challenging as the sampling probabilities in the random network are in fact a function of the input $\mathbf{x}+\boldsymbol{\delta}$ as well, making the overall minimization highly non-convex. In addition, there are several variations of sampling schemes based on different approximations of the output variance. Thus, it is difficult to analytically derive the sampling probabilities for various schemes and substitute them in \eqref{attack_final2}. In this work, we test the performance of the  proposed minimum-uncertainty based detection scheme against FGSM attacks targeting the objective in \eqref{attack_final2}, where the attack perturbation $\boldsymbol{\delta}$ is crafted as the stochastic gradient averaged over $R=50$ realization, defined as 
	\begin{equation}
	\boldsymbol{\delta} =  \epsilon \cdot \text{sign} ( \hat{\mathbf{g}})
	\end{equation}
	with 
	\begin{equation}\label{s-fgsm2}
	\hat{\mathbf{g}} := \nabla_{\mathbf{x}} \Big(\sum_{k=1}^K  y_{\nu,k}^{\text{target}} \log(y_{\nu,k}(\mathbf{x}))+  \dfrac{\mu}{R} \sum_{r=1}^R \|\mathbf{y}_{\nu}^{(r)}(\mathbf{x}) - \bar{ \mathbf{y}}_{\nu}(\mathbf{x})  \|_2^2\Big).\\
	\end{equation}
	Similar to Section 6.2, attacks are performed on the cats-and-dogs dataset, on clean images that were correctly classified by the deterministic network. Attack parameters are set as $\epsilon=10$ and $\epsilon=20$ for values of $\mu= 0.01, 0.1, 1$ with $R=50$, and the ROC curves are depicted in Fig. \ref{fig:roc-adaptive2}.

	Furthermore, attack success rate corresponding to the percentage of clean images that were correctly classified by the full network and misclassified if perturbed by \eqref{s-fgsm2}, as well as the AUC-ROC for $\epsilon=10$ and varying values of $\mu$ is reported in Fig. \ref{fig:auc_and_acc2}. It is interesting to note that in all detection mechanisms, incorporating the variance minimization lowers the attack success rate. However, in DVM-log sampling schemes, where the sampling probabilities are dynamically updated per layer effective in high $C$ regimes, provides the highest robustness in terms of attack success rate reduction as well as the AUC-ROC. Attack success rate also drops significantly in VM and VM-lin while the AUC-ROCs exhibit slight decrease. In contrast, attack success reduction occurs at higher $\mu$ values in Dropout and SAP detection schemes, accompanied with a slight increase in the AUC-ROC.}
	
\section{Conclusions and research outlook}\label{sec:conclusion}
Safe and reliable utilization of state-of-the-art CNNs is contingent upon robustifying their performance against adversarial perturbations. To this end, { and inspired by Bayesian neural networks}, we have investigated attack detection, where through imposing a certain level of randomness we designed the variational distribution to minimize network uncertainty. The premise is that the inherent distance of the adversarial perturbation from the natural-image manifold will cause the overall network uncertainty to exceed that of the clean image, and thus facilitate successful detection. Network uncertainty was expressed as a summation of its layer-wise components, whose  exact as well as approximate minimizers have been developed. Links with recent sampling-based approaches have been delineated, along with efficient implementations of the proposed approach. Finally, numerical tests on the CIFAR10 and the cats-and-dogs datasets on deep state-of-the-art CNNs demonstrated the importance of placement as well as tuning of the sampling parameters, which readily translate to improved attack detection. 

Among future directions, one can incorporate the novel approach in attack correction schemes based on randomization. This is of particular importance for careful sampling at the initial layers of the network. One can further exploit an ensemble of detection networks, in which sampling units are incorporated at \emph{random} depths. In addition to increased defense strength, the latter introduce a second source of randomness in the defense mechanism, and thus prevent identification by the attackers. {Provable detection with performance guarantees is also among future directions. }

\begin{table*}[!h]
	\begin{center}
		\small
		\begin{tabular}{ |c||c |c| c|c||c|c||c|c| } 
			\hline
			&\multicolumn{4}{c||}{ BIM Attack} & \multicolumn{2}{c||}{ C\&W Attack} & \multicolumn{2}{c|}{Combination  Attack}  \\
			
			Sampling &\multicolumn{2}{c|}{$\epsilon=10$}&\multicolumn{2}{c||}{$\epsilon=20$}& \multicolumn{2}{c||}{}&\multicolumn{2}{c|}{}\\ 
			\cline{2-9}
			
			Method& parameters& \small{AUC}  & parameters & AUC  & parameters & AUC  &  parameters & AUC  \\ 
			
			\hline
			
			{\footnotesize {VM}}   & \multirow{3}{4em}{}  & \bf 71.4 & 	& \bf  \bf 70.7 & &  71.6	&  & \bf 70.4 \\ 
			\cline{1-1}  \cline{3-3} \cline{5-5} \cline{7-7} \cline{9-9}
			
			\footnotesize VM-log&\small$(B,f)=(1,3.0)$& 70.3 & \small$(B,f)=(1,3.0)$& 69.4 &\small$(B,f)=(2,0.8)$& 56.9 & \small$(B,f)=(2,0.7)$ & 57.0\\ 			
			\cline{1-1}  \cline{3-3} \cline{5-5} \cline{7-7} \cline{9-9} 
			
			\footnotesize VM-linear &   & 66.7 & 	& 66.2 &  & \bf 71.9	&  & 68.2 \\ 
			\hline

			\footnotesize DVM-log & \small$(B,f)=(4,3.0)$ & 63.9 &  \small$(B,f)=(4,3.0)$& 63.6 &  \small$(B,f)=(4,3.0)$ & 69.6	& \small$(B,f)=(4,3.0)$ & 68.1\\ 
			\hline
			
			\footnotesize SAP& \small$(B,f)=(1,3.0)$ & 67.1 &\small$(B,f)=(1,3.0)$ & 67.1 & \small$(B,f)=(5,0.9)$ &	63.8 & \small$(B,f)=(6,1.0)$ & 67.1 \\ 
			\hline
			
			\footnotesize Dropout	& \small$(B,\pi_{\text{\tiny drp}})=(5,0.6)$ & 60.1  & \small$(B,\pi_{\text{\tiny drp}})=(6,0.1)$& 59.6 & \small$(B,\pi_{\text{\tiny drp}})=(6,0.5)$ & 63.1	&\small$(B,\pi_{\text{\tiny drp}})=(6,0.1)$  &  {66.1}  \\ 
			\hline
		\end{tabular}
			\label{tab:auc-cad2}
	\end{center}
	
	\caption{AUC-ROC of different attack-detection sampling schemes on cats-and-dogs test set against BIM, C\&W, and combination attacks. Higher values indicate better detection.} 
\end{table*}

\section{Appendix}
\subsection{Proof of Proposition 1 }\label{sec:app_proof}
Define $\mathbf{u}_\nu:= \mathbf{W}_2 \sigma ( \mathbf{W}_1 \mathbf{x}_\nu)$ and 
approximate it using the first-order Taylor expansion around $\bar{\mathbf{u}}_\nu:= \mathbb{E}_{q_\theta(\omega)}[\mathbf{u}_\nu]$, to arrive at 
\begin{equation}\label{l1}
\mathbf{y}_\nu \simeq \sigma_{\text{softmax}}(\bar{\mathbf{u}}_\nu) + \nabla \sigma_{\text{softmax}}(\mathbf{u})\Big|_{\mathbf{u} = \bar{\mathbf{u}}_\nu} (\mathbf{u}_\nu - \bar{\mathbf{u}}_\nu)
\end{equation}
which after taking expectation yields 
\begin{equation}\label{l2}
\mathbb{E}_{q_\theta(\omega)}[\mathbf{y}_\nu] \simeq \sigma_{\text{softmax}}(\bar{\mathbf{u}}_\nu)\;.
\end{equation}
Upon defining the matrix $\mathbf{H}_1 := \nabla \sigma_{\text{softmax}}(\mathbf{u})\Big|_{\mathbf{u} = \bar{\mathbf{u}}_\nu}$, and using \eqref{l1} and \eqref{l2}, we find 
$\mathbf{y}_\nu - 	\mathbb{E}_{q_\theta(\omega)}[\mathbf{y}_\nu]  \simeq \mathbf{H}_1  (\mathbf{u}_\nu - \bar{\mathbf{u}}_\nu)
$
that leads to approximating the variance score as 
\begin{align*}
&\text{Cov}_{q_\theta(\omega)}[\mathbf{ y}_\nu] \\ &   =	\mathbb{E }_{q_\theta(\omega)} \Big[(\mathbf{y}_\nu - 	\mathbb{E}_{q_\theta(\omega)}[\mathbf{y}_\nu] ) (\mathbf{y}_\nu - 	\mathbb{E}_{q_\theta(\omega)}[\mathbf{y}_\nu] )^\top \Big] \\
&  	\simeq 	\mathbb{E }_{q_\theta(\omega)} \Big[ \mathbf{H}_1  (\mathbf{u}_\nu - \bar{\mathbf{u}}_\nu)(\mathbf{u}_\nu - \bar{\mathbf{u}}_\nu )^\top   \mathbf{H}_1^\top \Big] \;.
\end{align*}
The trace of the latter can be upper bounded by 
\begin{align*}
& \text{Tr (Cov}_{q_\theta(\omega)}[\mathbf{ y}_\nu] ) \leq \lambda_1  \text{Tr (Cov}_{q_\theta(\omega)}[\mathbf{u}_\nu] )   
\end{align*}
where $\text{Tr (Cov}_{q_\theta(\omega)}[\mathbf{ u}_\nu] )  :=  \text{Tr}(\mathbb{E}_{q_\theta(\omega)} [ (\mathbf{u}_\nu - \bar{\mathbf{u}}_\nu)(\mathbf{u}_\nu - \bar{\mathbf{u}}_\nu )^\top  ]) $, and 
$\lambda_1 := \text{Tr}(\mathbf{H}_1^\top \mathbf{H}_1) $ is \textit{deterministic}. 
Thus, the output variance score is upperbounded by that of the previous layer up to a constant $\lambda_1$. Repeating this process of approximating $\mathbf{u}_\nu$ as a function of $\mathbf{v}_\nu = \mathbf{W}_1 \mathbf{x}_\nu$ by the first-order Taylor expansion around $\bar{\mathbf{v}}_\nu: = \mathbb{E}_{\mathbf{W}_1} [\mathbf{v}_\nu] $, leads with
$\mathbf{H}_2:= \nabla \sigma(\bar{\mathbf{v}}_\nu) $ to $\mathbf{u}_\nu$ and its mean compensated approximation
\begin{align}\label{u_approximation}
\mathbf{u}_\nu &\simeq  \mathbf{W}_2 \sigma(\bar{\mathbf{v}}_\nu) + \mathbf{W}_2 \mathbf{H}_2  ( {\mathbf{v}}_\nu - \bar{\mathbf{v}}_\nu )\\ \nonumber
\bar{\mathbf{u}}_\nu & \simeq  \mathbb{E}_{\mathbf{W}_2}[\mathbf{W}_2 \sigma(\bar{\mathbf{v}}_\nu)] \\ \nonumber
\mathbf{u}_\nu - \bar{\mathbf{u}}_\nu  & \simeq  \mathbb{E}_{\mathbf{W}_2}[\mathbf{W}_2 \sigma(\bar{\mathbf{v}}_\nu)] - \mathbf{W}_2 \sigma(\bar{\mathbf{v}}_\nu) + \mathbf{W}_2 \mathbf{H}_2  ( {\mathbf{v}}_\nu - \bar{\mathbf{v}}_\nu )\:. 
\end{align}
The latter yields the covariance approximation 
\begin{align*}
&	\text{ Cov}_{q_\theta(\omega)}[\mathbf{ u}_\nu]  \simeq 
\text{ Cov}_{\mathbf{W}_2}[\mathbf{W}_2 \sigma(\bar{\mathbf{v}}_\nu) ] \\&
+  \mathbb{E}_{\mathbf{W}_2}[\mathbf{W}_2 \mathbf{H}_2  \mathbb{E}_{\mathbf{W}_1}[( {\mathbf{v}}_\nu - \bar{\mathbf{v}}_\nu ) ( {\mathbf{v}}_\nu - \bar{\mathbf{v}}_\nu )^\top] \mathbf{H}_2^\top  \mathbf{W}^\top_2 ] \\
& + \mathbb{E}_{\mathbf{W}_2}\Big[   \mathbf{W}_2 \mathbf{H}_2  \mathbb{E}_{\mathbf{W}_1}( {\mathbf{v}}_\nu - \bar{\mathbf{v}}_\nu) \\ &  \quad \qquad \quad \times \Big(\mathbb{E}_{\mathbf{W}_2}[\mathbf{W}_2 \sigma(\bar{\mathbf{v}}_\nu)] - \mathbf{W}_2 \sigma(\bar{\mathbf{v}}_\nu)\Big)^\top   \Big]\\
& = 	\text{Cov}_{\mathbf{W}_2}[\mathbf{W}_2 \sigma(\bar{\mathbf{v}}_\nu) ] +  \mathbb{E}_{\mathbf{W}_2}[\mathbf{W}_2 \mathbf{H}_2 \text{ Cov}_{\mathbf{W}_1}[{\mathbf{v}}_\nu ]   \mathbf{H}_2^\top  \mathbf{W}^\top_2 ]
\end{align*}
where we have used the independence of random matrices $\mathbf{W}_1$ and $\mathbf{W}_2$, and $\mathbb{E}_{\mathbf{W}_1}( {\mathbf{v}}_\nu - \bar{\mathbf{v}}_\nu)= \bf 0$. Taking the trace and using the inequality $\text{Tr}(\mathbf{AB})\leq \text{Tr}(\mathbf{A})\text{Tr}(\mathbf{B})$ for positive semi-definite matrices $\mathbf{A},\mathbf{B} \geq \mathbf{0}$ twice, we arrive after defining   $\lambda_2:=\text{Tr}(\mathbf{H}_2^\top \mathbf{H}_2)$, at
\begin{align*}
&\text{Tr}(	\text{ Cov}_{q_\theta(\omega)}[\mathbf{ u}_\nu] ) \\
& \simeq  	\text{Cov}_{\mathbf{W}_2}[\mathbf{W}_2 \sigma(\bar{\mathbf{v}}_\nu) ] )  	+  \mathbb{E}_{\mathbf{W}_2}[ \text{Tr}(\mathbf{W}_2 \mathbf{H}_2 \text{ Cov}_{\mathbf{W}_1}[{\mathbf{v}}_\nu ]   \mathbf{H}_2^\top  \mathbf{W}^\top_2) ]  \\
&\leq 	\text{Tr} (\text{Cov}_{\mathbf{W}_2}[\mathbf{W}_2 \sigma(\bar{\mathbf{v}}_\nu) ] )    \\
& \quad  + \lambda_2 \text{Tr}( \mathbb{E}_{\mathbf{W}_2}[ \mathbf{W}_2    \mathbf{W}^\top_2]  ) \text{Tr}(\text{ Cov}_{\mathbf{W}_1}[{\mathbf{v}}_\nu ]  )\;.
\end{align*}
Leveraging the last inequality, we can majorize the uncertainty minimization in \eqref{opt_var} by that in \eqref{opt_majorized}. This is a coupled minimization of layer-wise variance scores $\text{Tr}(\text{ Cov}_{\mathbf{W}_1}[{\mathbf{v}}_\nu ]  ) $ and $\text{Tr} (\text{Cov}_{\mathbf{W}_2}[\mathbf{W}_2 \sigma(\bar{\mathbf{v}}_\nu) ] )$, that we solve as follows. 

Using $\mathbf{W}_1^{(TR)}$ along with \eqref{W} and \eqref{M2}, we have $\mathbf{W}_1=$ $ \mathbf{W}_1^{(TR)} \mathbf{S}_1 \mathbf{D}_1$, where $\mathbf{S}_1:= \text{diag}( [z_{1,1}, z_{1,2}, \cdots, z_{1,h_1}] )$ is the sampling matrix with its pseudo-inverse diagonal mean  
$\mathbf{D}_1 := \text{diag}^{\dagger}\Big(\mathbb{E}_{q(\mathbf{z}; \mathbf{p}_1)} [z_{1,1}, z_{1,2}, \ldots, z_{1,h_1}] \Big)$. This implies that 
$\bar{\mathbf{v}}_\nu : = \mathbb{E}_{\mathbf{W}_1} [\mathbf{v}_\nu] = \mathbf{W}_1^{(TR)} \mathbf{x}_\nu$, which does not depend on the sampling vector $\mathbf{p}_1$. As a result, the minimization in \eqref{opt_majorized} can be readily solved by the proposed subproblems. $\blacksquare$

\begin{figure*}
	\centering
	\includegraphics[scale=0.33]{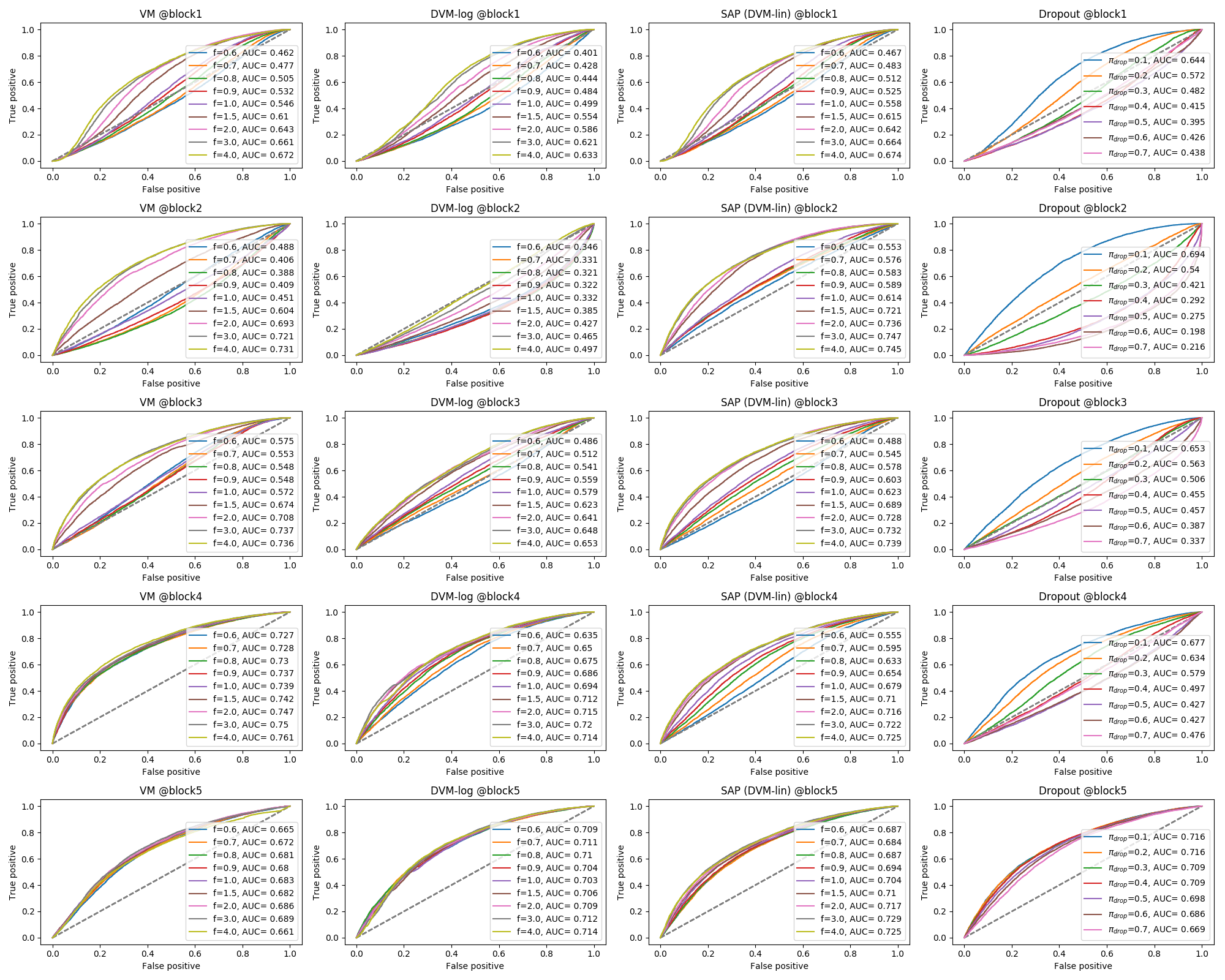}
				\vspace{-0.4cm}
	\caption{Performance of different sampling mechanisms at various depths and parameters against combination attack on the CIFAR10 dataset}
	\label{fig:sensitivity-cifar}
			\vspace{-0.4cm}
\end{figure*}

\begin{figure*}
	\centering
	\includegraphics[scale=0.33]{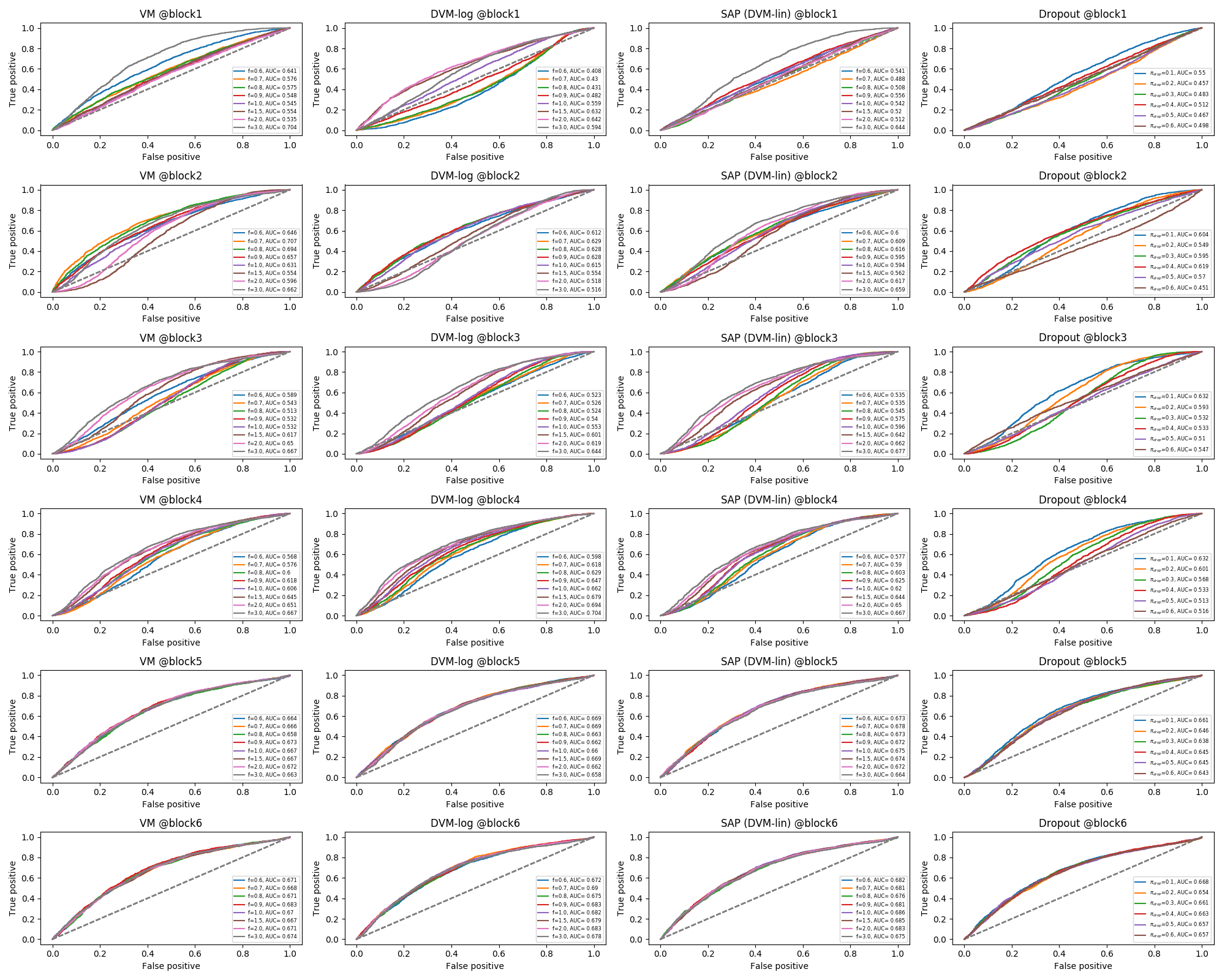}
				\vspace{-0.4cm}
	\caption{Performance of different sampling mechanisms at various depths and parameters against combination attack on the cats-and-dogs dataset}
	\label{fig:sensitivity-cad}
			\vspace{-0.4cm}
\end{figure*}

\subsection{Proof of Proposition 2}\label{sec:app_proof2}
To solve \eqref{opt_general}, consider the Lagrangian
\begin{equation*}\label{L}
L = \sum_{i=1}^h \dfrac{\alpha_i}{1-e^{-Cp_i} } + \rho (\mathbf{1}^\top \mathbf{p}-1)  \qquad 0 \leq p_i \leq 1
\end{equation*}
and upon setting its gradient to zero
\begin{equation}\label{grad_L}
\dfrac{\partial L}{\partial p_i} = \dfrac{-C \alpha_i e^{-C p_i}}{(1-e^{-Cp_i})^2} +\rho = 0
\end{equation}
and introducing the change of variable 
\begin{equation*}
y_i := exp(-Cp_i) \quad e^{-C} < y_i <1
\end{equation*}
we find that \eqref{grad_L} reduces to $\rho' y_i^2 - (2\rho'+\alpha_i) y_i + \rho' = 0\;	.$
The feasible root of this quadratic polynomial is 
\begin{equation*}
y_i = \dfrac{2 \rho' + \alpha_i  - \sqrt{(2\rho'+\alpha_i)^2-4\rho'^2} }{2\rho'}\;.
\end{equation*}
Using the simplex constraint at the optimal point, we find 
\begin{equation*}
-\dfrac{1}{C}\sum_i \ln y_i = 1 
\end{equation*}
which after reverting the change of variable, reduces the optimization in \eqref{opt_general} to the following root-finding task
\begin{equation*}
\sum_i \ln (2 \rho' + \alpha_i  - \sqrt{(2\rho'+\alpha_i)^2-4\rho'^2}) -n \ln (2\rho') +C = 0 .
\end{equation*}
This scalar root-finding problem can be solved using bisection that enjoys super-linear convergence rate. $\blacksquare$
\subsection{Selection of defense parameters}\label{sec:app_sensitivity}
In order to further provide insight on the performance against various selection of $B$, $f$, and $\pi$ parameters, Figs. \ref{fig:sensitivity-cifar} and \ref{fig:sensitivity-cad} illustrate the AUC-ROC for VM-exact, DVM-log, SAP, and uniform dropout against the combination attack. As the plots suggest, uniform dropout reaches its best performance when placed at the last block, whereas higher performance can be obtained by placing carefully-tuned sampling at units in hidden layers before the last. Furthermore, at a given block $B$, VM-exact demonstrates higher robustness for different values of $f$; that is, smaller fluctuation in AUC is observed, whereas other methods are usually more prone to under-performance given sub-optimal parameters. 

\bibliographystyle{IEEEtran}
\bibliography{adv_tpami} 

\begin{thebibliography}{10}
\providecommand{\url}[1]{#1}
\csname url@samestyle\endcsname
\providecommand{\newblock}{\relax}
\providecommand{\bibinfo}[2]{#2}
\providecommand{\BIBentrySTDinterwordspacing}{\spaceskip=0pt\relax}
\providecommand{\BIBentryALTinterwordstretchfactor}{4}
\providecommand{\BIBentryALTinterwordspacing}{\spaceskip=\fontdimen2\font plus
\BIBentryALTinterwordstretchfactor\fontdimen3\font minus
  \fontdimen4\font\relax}
\providecommand{\BIBforeignlanguage}[2]{{%
\expandafter\ifx\csname l@#1\endcsname\relax
\typeout{** WARNING: IEEEtran.bst: No hyphenation pattern has been}%
\typeout{** loaded for the language `#1'. Using the pattern for}%
\typeout{** the default language instead.}%
\else
\language=\csname l@#1\endcsname
\fi
#2}}
\providecommand{\BIBdecl}{\relax}
\BIBdecl

\bibitem{simonyan2014very}
K.~Simonyan and A.~Zisserman, ``Very deep convolutional networks for
  large-scale image recognition,'' in \emph{Intl. Conf. on Learning
  Representations}, San Diego, CA, USA, May 2015.

\bibitem{redmon2017yolo9000}
J.~Redmon and A.~Farhadi, ``Yolo9000: Better, faster, stronger,'' in \emph{IEEE
  Conf. on Computer Vision and Pattern Recognition}, Honolulu, HI, USA, June
  2017, pp. 6517--6525.

\bibitem{sharif2016accessorize}
M.~Sharif, S.~Bhagavatula, L.~Bauer, and M.~K. Reiter, ``Accessorize to a
  crime: Real and stealthy attacks on state-of-the-art face recognition,'' in
  \emph{Proc. of Conf. on Computer and Communications Security}, Vienna,
  Austria, Oct. 2016, pp. 1528--1540.

\bibitem{sutskever2014sequence}
I.~Sutskever, O.~Vinyals, and Q.~V. Le, ``Sequence to sequence learning with
  neural networks,'' in \emph{Advances in Neural Information Processing
  Systems}, Montréal, Canada, Dec. 2014, pp. 3104--3112.

\bibitem{engel2017neural}
J.~Engel, C.~Resnick, A.~Roberts, S.~Dieleman, M.~Norouzi, D.~Eck, and
  K.~Simonyan, ``Neural audio synthesis of musical notes with wavenet
  autoencoders,'' in \emph{Intl. Conf. on Machine Learning}, Sydney, Australia,
  May 2017, pp. 1068--1077.

\bibitem{szegedy2013intriguing}
C.~Szegedy, W.~Zaremba, I.~Sutskever, J.~Bruna, D.~Erhan, I.~Goodfellow, and
  R.~Fergus, ``Intriguing properties of neural networks,'' \emph{Intl. Conf. of
  Learning Representations}, Banff, Canada, May 2014.

\bibitem{moosavi2017universal}
S.~M. Moosavi-Dezfooli, A.~Fawzi, O.~Fawzi, and P.~Frossard, ``Universal
  adversarial perturbations,'' in \emph{IEEE Conf. on Computer Vision and
  Pattern Recognition}, Honolulu, HI, USA, June 2017, pp. 1765--1773.

\bibitem{tsipras2018there}
D.~Tsipras, S.~Santurkar, L.~Engstrom, A.~Turner, and A.~Madry, ``Robustness
  may be at odds with accuracy,'' \emph{Intl. Conf. on Learning
  Representation}, New Orleans, La, USA, May 2019.

\bibitem{eykholt2018robust}
K.~Eykholt, I.~Evtimov, E.~Fernandes, B.~Li, A.~Rahmati, C.~Xiao, A.~Prakash,
  T.~Kohno, and D.~Song, ``Robust physical-world attacks on deep learning
  visual classification,'' in \emph{IEEE Conf. on Computer Vision and Pattern
  Recognition}, Salt Lake City, Utah, USA, June 2018, pp. 1625--1634.

\bibitem{zhang2017dolphinattack}
G.~Zhang, C.~Yan, X.~Ji, T.~Zhang, T.~Zhang, and W.~Xu, ``Dolphinattack:
  Inaudible voice commands,'' in \emph{ACM Conf. on Computer and Communications
  Security}, Dallas, TX, USA, Oct. 2017, pp. 103--117.

\bibitem{goswami2018unravelling}
G.~Goswami, N.~Ratha, A.~Agarwal, R.~Singh, and M.~Vatsa, ``Unravelling
  robustness of deep learning based face recognition against adversarial
  attacks,'' \emph{AAAI Conf. on Artificial Intelligence}, New Orleans, LA,
  USA, Feb. 2018.

\bibitem{bose2018adversarial}
A.~J. Bose and P.~Aarabi, ``Adversarial attacks on face detectors using neural
  net based constrained optimization,'' \emph{IEEE Intl. Workshop on Multimedia
  Signal Processing}, Vancouver, Canada, Aug. 2018.

\bibitem{carlini2017adversarial}
N.~Carlini and D.~Wagner, ``Adversarial examples are not easily detected:
  Bypassing ten detection methods,'' in \emph{ACM Workshop on Artificial
  Intelligence and Security}, Dallas, TX, USA, Oct. 2017, pp. 3--14.

\bibitem{carlini2017towards}
------, ``Towards evaluating the robustness of neural networks,'' in \emph{IEEE
  Symposium on Security and Privacy}, San Jose, CA, USA, May 2017, pp. 39--57.

\bibitem{papernot2016transferability}
N.~Papernot, P.~McDaniel, and I.~Goodfellow, ``Transferability in machine
  learning: From phenomena to black-box attacks using adversarial samples,''
  \emph{Technical Report, arXiv preprint arXiv:1605.07277}, 2016.

\bibitem{zhang2019limitations}
H.~Zhang, H.~Chen, Z.~Song, D.~Boning, I.~S. Dhillon, and C.-J. Hsieh, ``The
  limitations of adversarial training and the blind-spot attack,'' \emph{Intl.
  Conf. on Learning Representations}, New Orleans , LA, USA, May 2019.

\bibitem{hong2019zero}
S.~Liu, P.-Y. Chen, X.~Chen, and M.~Hong, ``Signsgd via zeroth-order oracle,''
  \emph{Intl. Conf. on Learning Representations}, New Orleans , LA, USA, May
  2019.

\bibitem{papernot2016limitations}
N.~Papernot, P.~McDaniel, S.~Jha, M.~Fredrikson, Z.~B. Celik, and A.~Swami,
  ``The limitations of deep learning in adversarial settings,'' in \emph{IEEE
  European Symposium on Security and Privacy}, Saarbrücken, Germany, March
  2016, pp. 372--387.

\bibitem{gu2014towards}
S.~Gu and L.~Rigazio, ``Towards deep neural network architectures robust to
  adversarial examples,'' \emph{Intl. Conf. on Learning Representations}, San
  Diego, CA, USA, May 2015.

\bibitem{metzen2017detecting}
J.~H. Metzen, T.~Genewein, V.~Fischer, and B.~Bischoff, ``On detecting
  adversarial perturbations,'' \emph{Intl. Conf. on Learning Representations},
  Toulon, France, April 2017.

\bibitem{lu2017safetynet}
J.~Lu, T.~Issaranon, and D.~Forsyth, ``Safetynet: Detecting and rejecting
  adversarial examples robustly,'' in \emph{IEEE Intl. Conf. on Computer
  Vision}, Venice, Italy, Oct. 2017, pp. 446--454.

\bibitem{feinman2017detecting}
R.~Feinman, R.~R. Curtin, S.~Shintre, and A.~B. Gardner, ``Detecting
  adversarial samples from artifacts,'' \emph{arXiv preprint arXiv:1703.00410},
  2017.

\bibitem{random_certified}
J.~M. Cohen, E.~Rosenfeld, and J.~Z. Kolter, ``Certified adversarial robustness
  via randomized smoothing,'' in \emph{Intl' Conf. on Machinbe Learning
  (ICML)}, Long Beach, CA, June 2019.

\bibitem{guo2017countering}
C.~Guo, M.~Rana, M.~Cisse, and L.~Van Der~Maaten, ``Countering adversarial
  images using input transformations,'' \emph{Intl. Conf. on Learning
  Representations}, Vancouver, Canada, May 2018.

\bibitem{gopalakrishnan2018combating}
S.~Gopalakrishnan, Z.~Marzi, U.~Madhow, and R.~Pedarsani, ``Combating
  adversarial attacks using sparse representations,'' \emph{arXiv preprint
  arXiv:1803.03880}, 2018.

\bibitem{miyato2018virtual}
T.~Miyato, S.~Maeda, S.~Ishii, and M.~Koyama, ``Virtual adversarial training: A
  regularization method for supervised and semi-supervised learning,''
  \emph{IEEE Trans. on Pattern Analysis and Machine Intelligence}, 2018.

\bibitem{schmidt2018adversarially}
L.~Schmidt, S.~Santurkar, D.~Tsipras, K.~Talwar, and A.~Madry, ``Adversarially
  robust generalization requires more data,'' \emph{Advances in Neural
  Information Processing Systems}, Montreal, Canada, Dec. 2018.

\bibitem{sinha2017certifying}
A.~Sinha, H.~Namkoong, and J.~Duchi, ``Certifying some distributional
  robustness with principled adversarial training,'' \emph{Intl. Conf of
  Learning Representations}, Vancouver, Canada, April 2018.

\bibitem{EricZico}
E.~Wong and J.~Z. Kolter, ``Provable defenses against adversarial examples via
  the convex outer adversarial polytope,'' in \emph{Intl. Conf. on Machinbe
  Learning (ICML)}, Stockholm, Sweden, June 2018.

\bibitem{towards_2019}
H.~Zhang, H.~Chen, C.~Xiao, B.~Li, D.~Boning, and C.-J. Hsieh, ``Towards stable
  and efficient training of verifiably robust neural networks,'' \emph{arXiv
  preprint arXiv:1906.06316}, 2019.

\bibitem{NIPS2018_sparse}
Y.~Guo, C.~Zhang, C.~Zhang, and Y.~Chen, ``Sparse {D}{N}{N}s with improved
  adversarial robustness,'' in \emph{Advances in Neural Information Processing
  Systems}, Montreal, Canada, Dec. 2018, pp. 240--249.

\bibitem{targetted_dropout}
A.~N. Gomez, I.~Zhang, K.~Swersky, Y.~Gal, and G.~E. Hinton, ``Targetted
  dropout,'' in \emph{Advances in Neural Information Processing Workshop
  Track}, Montreal, Canada, USA, Dec. 2018.

\bibitem{zhang2016understanding}
C.~Zhang, S.~Bengio, M.~Hardt, B.~Recht, and O.~Vinyals, ``Understanding deep
  learning requires rethinking generalization,'' \emph{Intl. Conf. on Learning
  Representations}, Toulon, France, April 2017.

\bibitem{gouk2018regularisation}
H.~Gouk, E.~Frank, B.~Pfahringer, and M.~Cree, ``Regularisation of neural
  networks by enforcing {L}ipschitz continuity,'' \emph{arXiv preprint
  arXiv:1804.04368}, 2018.

\bibitem{hendrycks17baseline}
D.~Hendrycks and K.~Gimpel, ``A baseline for detecting misclassified and
  out-of-distribution examples in neural networks,'' \emph{Proc. of Intl. Conf.
  on Learning Representations}, Toulon, France, May 2017.

\bibitem{calibration}
C.~Guo, G.~Pleiss, Y.~Sun, and K.~Q. Weinberger, ``On calibration of modern
  neural networks,'' pp. 1321--1330, Sydney, Australia, Aug. 2017.

\bibitem{malinin2018predictive}
A.~Malinin and M.~Gales, ``Predictive uncertainty estimation via prior
  networks,'' in \emph{Advances in Neural Information Processing Systems},
  Montreal, Canada, Dec. 2018, pp. 7047--7058.

\bibitem{smith2018understanding}
L.~Smith and Y.~Gal, ``Understanding measures of uncertainty for adversarial
  example detection,'' \emph{Conf. on Uncertainty in Artificial Intelligence},
  Monterey, CA, USA, Aug. 2018.

\bibitem{kendall2017uncertainties}
A.~Kendall and Y.~Gal, ``What uncertainties do we need in bayesian deep
  learning for computer vision?'' in \emph{Advances in Neural Information
  Processing Systems}, Long Beach, CA, Dec. 2017, pp. 5574--5584.

\bibitem{icassp}
F.~Sheikholeslami, S.~Jain, and G.~B. Giannakis, ``Efficient randomized defense
  against adversarial attacks in deep convolutional networks,'' in \emph{Proc.
  of Intl. Conf. on Acoustics, Speech, and Signal Processing}, Brighton, UK,
  May 12-17, 2019.

\bibitem{gal2016bayesian}
Y.~Gal and Z.~Ghahramani, ``Bayesian convolutional neural networks with
  {B}ernoulli approximate variational inference,'' in \emph{Intl. Conf. on
  Learning Representations Workshop Track}, San Juan, Puerto Rico, May 2016.

\bibitem{gal2016theoretically}
------, ``A theoretically grounded application of dropout in recurrent neural
  networks,'' in \emph{Advances in Neural Information Processing Systems},
  Barcelona, Spain, Dec. 2016, pp. 1019--1027.

\bibitem{Statistical_2019}
L.~Cardelli, M.~Kwiatkowska, L.~Laurenti, N.~Paoletti, A.~Patane, and
  M.~Wicker, ``Statistical guarantees for the robustness of {B}ayesian neural
  networks,'' \emph{arXiv preprint arXiv:1903.01980}, 2019.

\bibitem{dhillon2018stochastic}
G.~S. Dhillon, K.~Azizzadenesheli, Z.~C. Lipton, J.~Bernstein, J.~Kossaifi,
  A.~Khanna, and A.~Anandkumar, ``Stochastic activation pruning for robust
  adversarial defense,'' \emph{Intl. Conf. on Learning Representationa},
  Vancouver, Canada, April 2018.

\bibitem{srivastava2014dropout}
N.~Srivastava, G.~Hinton, A.~Krizhevsky, I.~Sutskever, and R.~Salakhutdinov,
  ``Dropout: A simple way to prevent neural networks from overfitting,''
  \emph{Journal of Machine Learning Research}, vol.~15, no.~1, pp. 1929--1958,
  2014.

\bibitem{bishop}
C.~Bishop, \emph{Pattern Recognition and Machine Learning}, Springer, New York,
  USA, 2006.

\bibitem{neal2012bayesian}
R.~M. Neal, \emph{Bayesian Learning for Neural Networks}.\hskip 1em plus 0.5em
  minus 0.4em\relax Springer Science \& Business Media, New York, USA, 2012.

\bibitem{garcia2018behavior}
D.~Garcia-Gasulla, F.~Pares, A.~Vilalta, J.~Moreno, E.~Ayguade, J.~Labarta,
  U.~Cortes, and T.~Suzumura, ``On the behavior of convolutional nets for
  feature extraction,'' \emph{Journal of Artificial Intelligence Research},
  vol.~61, pp. 563--592, 2018.

\bibitem{he2016deep}
K.~He, X.~Zhang, S.~Ren, and J.~Sun, ``Deep residual learning for image
  recognition,'' in \emph{IEEE Conf. on Computer Vision and Pattern
  Recognition}, Las Vegas Valley, NV, USA, June 2016, pp. 770--778.

\bibitem{goodfellow2014explaining}
I.~J. Goodfellow, J.~Shlens, and C.~Szegedy, ``Explaining and harnessing
  adversarial examples,'' \emph{Intl. Conf. on Learning Representations}, San
  Diego, CA, USA, May 2015.

\bibitem{kurakin2016adversarial}
A.~Kurakin, I.~Goodfellow, and S.~Bengio, ``Adversarial examples in the
  physical world,'' \emph{Intl. Conf. on Learning Representations}, Toulon,
  France, April 2017.

\bibitem{dong2018boosting}
Y.~Dong, F.~Liao, T.~Pang, H.~Su, J.~Zhu, X.~Hu, and J.~Li, ``Boosting
  adversarial attacks with momentum,'' in \emph{Proc. of IEEE Conf. on Computer
  Vision and Pattern Recognition}, Salt Lake City, UT, USA, June 2018, pp.
  9185--9193.

\end{thebibliography}

\begin{IEEEbiography}[{\includegraphics[width = 1in,height=1.25in,clip,keepaspectratio]{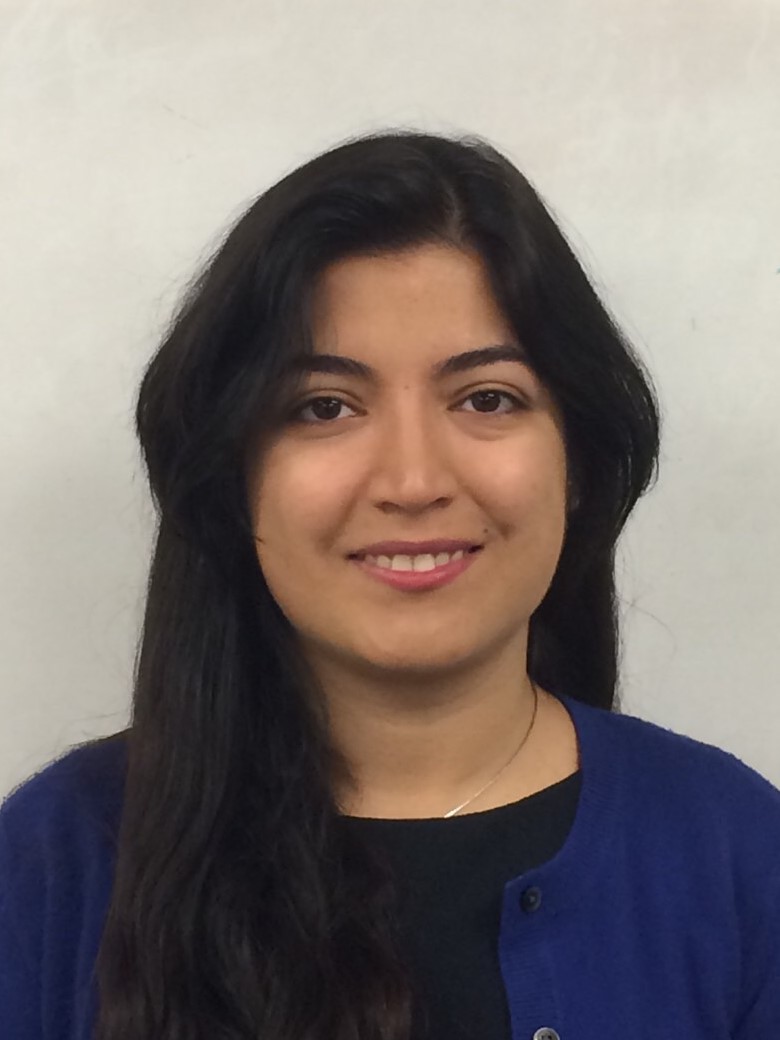}}]{Fatemeh Sheikholeslami} (S'13) received the B.Sc. and M.Sc. degrees in Electrical Engineering from the University of Tehran and Sharif University of Technology, Tehran, Iran, in 2010 and 2012, respectively. Since September 2013, she has been working toward the Ph.D. degree with the Department of Electrical and Computer Engineering, University of Minnesota, MN, USA. Her research interests include Machine Learning and Network Science.
\end{IEEEbiography}

\begin{IEEEbiography}[{\includegraphics[width = 1in,height=1.25in,clip,keepaspectratio]{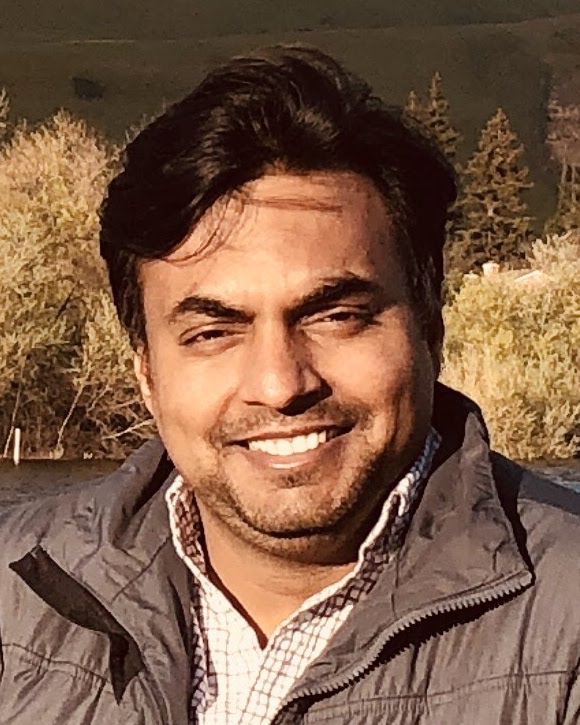}}]{Swayambhoo Jain} (S'13)  received the Bachelor of Technology degree in Electronics and Communication Engineering from the National Institute of Technology, Calicut, India in 2007. He was a Research and Development Engineer with IBM Systems and Technology Group, Bangalore, India, from 2007 to 2010. He received the M.Sc. degree  and Ph.D. degree in Electrical Engineering from the University of Minnesota, MN, USA, in 2012 and 2017, respectively. Since 2017 he has been working as a Researcher at Technicolor Artificial Intelligence Lab in Palo Alto, CA, USA. His general research interests lie in Machine Learning, Signal Processing, and High Dimensional Statistical Inference. His current research focus in on issues related to efficiency and robustness in Deep Learning. 
\end{IEEEbiography}

 \begin{IEEEbiography}[{\includegraphics[width=1in,height=1.25in,clip,keepaspectratio]{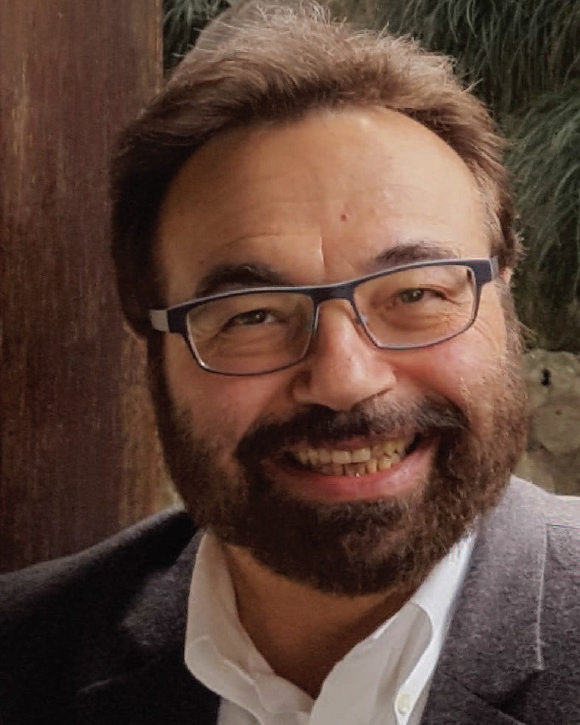}}]{G. B.  Giannakis} (Fellow'97) received his Diploma in Electrical
	Engr. from the Ntl. Tech. Univ. of Athens, Greece, 1981. From
	1982 to 1986 he was with the Univ. of Southern California (USC),
	where he received his MSc. in Electrical Engineering, 1983, MSc.
	in Mathematics, 1986, and Ph.D. in Electrical Engr., 1986. He
	was with the University of Virginia from 1987 to 1998, and since
	1999 he has been a professor with the Univ. of Minnesota, where
	he holds an Endowed Chair in Wireless Telecommunications, a
	University of Minnesota McKnight Presidential Chair in ECE,
	and serves as director of the Digital Technology Center.
	
	His general interests span the areas of communications, networking
	and statistical learning - subjects on which he has published
	more than 450 journal papers, 750 conference papers, 25 book
	chapters, two edited books and two research monographs (h-index 140).
	Current research focuses on Data Science, Internet of Things, and
	Network Science with applications to social, brain, and power
	networks with renewables. He is the (co-) inventor of 32 patents issued,
	and the (co-) recipient of 9 best journal paper awards from the
	IEEE Signal Processing (SP) and Communications Societies, including
	the G. Marconi Prize Paper Award in Wireless Communications. He also
	received Technical Achievement Awards from the SP Society (2000),
	from EURASIP (2005), a Young Faculty Teaching Award, the G. W. Taylor
	Award for Distinguished Research from the University of Minnesota,
	and the IEEE Fourier Technical Field Award (inaugural recipient in
	2015). He is a Fellow of EURASIP, and has served the IEEE in a number
	of posts, including that of a Distinguished Lecturer for the IEEE-SPS.
\end{IEEEbiography}
\end{document}